\documentclass[11pt,a4paper]{article}
\usepackage[hyperref]{blackbox}
\usepackage{times}
\usepackage{latexsym}

\usepackage{microtype}

\usepackage{times}
\usepackage{microtype}
\usepackage{graphicx}
\usepackage{subcaption}
\usepackage{booktabs} 
\usepackage{xcolor}
\usepackage{amsfonts}
\usepackage{nicefrac}
\usepackage{float}
\usepackage[most]{tcolorbox}
\graphicspath{ {./images/} }

\aclfinalcopy 

\title{Discovering the Compositional Structure of Vector Representations with Role Learning Networks}

\author{First Author \\
  Affiliation / Address line 1 \\
  Affiliation / Address line 2 \\
  Affiliation / Address line 3 \\
  \texttt{email@domain} \\\And
  Second Author \\
  Affiliation / Address line 1 \\
  Affiliation / Address line 2 \\
  Affiliation / Address line 3 \\
  \texttt{email@domain} \\\And
  Third Author \\
  Affiliation / Address line 1 \\
  Affiliation / Address line 2 \\
  Affiliation / Address line 3 \\
  \texttt{email@domain} \\\\\And
  Paul Smolensky \\
   \\
  Redmond, WA 98052 \& \\
  Johns Hopkins University \\
  \texttt{} \\
  \\}
  
\author{Paul Soulos,\textsuperscript{1} R. Thomas McCoy,\textsuperscript{1} Tal Linzen,\textsuperscript{2} and Paul Smolensky\textsuperscript{3,1} \\
\textsuperscript{1}Department of Cognitive Science, Johns Hopkins University \\
\textsuperscript{2}Department of Linguistics and Center for Data Science, New York University \\
\textsuperscript{3}Microsoft Research \\
\texttt{psoulos1@jhu.edu},\texttt{tom.mccoy@jhu.edu}, \texttt{linzen@nyu.edu}, \texttt{psmo@microsoft.com} \\}

\date{}

\usepackage{amsmath,amsfonts,bm}

\newcommand{\newterm}[1]{{\bf #1}}

\def\eqref#1{equation~\ref{#1}}

\def\1{\bm{1}}

\def\va{{\bm{a}}}
\def\vb{{\bm{b}}}

\def\vf{{\bm{f}}}

\def\vk{{\bm{k}}}

\def\vm{{\bm{m}}}

\def\vq{{\bm{q}}}
\def\vr{{\bm{r}}}
\def\vs{{\bm{s}}}

\def\vv{{\bm{v}}}

\def\mA{{\bm{A}}}

\def\mF{{\bm{F}}}

\def\mR{{\bm{R}}}

\DeclareMathAlphabet{\mathsfit}{\encodingdefault}{\sfdefault}{m}{sl}
\SetMathAlphabet{\mathsfit}{bold}{\encodingdefault}{\sfdefault}{bx}{n}

\newcommand{\R}{\mathbb{R}}

\newcommand{\softmax}{\mathrm{softmax}}

\DeclareMathOperator*{\argmax}{arg\,max}

\usepackage{color}
\usepackage{caption}
\mathchardef\mhyphen="2D
\newcommand{\hs}{\hspace{1mm}}

\renewcommand{\t}[1]{{\mathtt{#1}}}
\renewcommand{\r}[1]{{\mathrm{#1}}}
\renewcommand{\b}[1]{{\mathbf{#1}}}
\renewcommand{\c}[1]{{\mathcal{#1}}}
\newcommand{\rR}{\r{R}}
\newcommand{\rF}{\r{F}}
\newcommand{\bW}{{\mathbb{W}}}

\newcommand{\RLN}{ROLE}

\newcommand{\hi}[1]{{\color{blue} #1}}

\definecolor{mygray}{RGB}{201, 50, 40}

\usepackage{soul}

\begin{document}
\maketitle
\begin{abstract}
How can neural networks perform so well on compositional tasks even though they lack explicit compositional representations?
We use a novel analysis technique called ROLE to show that recurrent neural networks perform well on such tasks by converging to solutions which implicitly represent symbolic structure. This method uncovers a symbolic structure which, when properly embedded in vector space, closely approximates the encodings of a standard seq2seq network trained to perform the compositional SCAN task. We verify the causal importance of the discovered symbolic structure by showing that, when we systematically manipulate hidden embeddings based on this symbolic structure, the model's output is changed in the way predicted by our analysis.
\end{abstract}

\section{Introduction} \label{sec:Intro}

\begin{figure}
\footnotesize

\definecolor{myblue}{RGB}{255,255,255}
\definecolor{mybluebox}{RGB}{64,64,64}

\begin{tcolorbox}[width=\columnwidth, colback={myblue}, coltitle=white, toptitle=3pt, bottomtitle=3pt, colframe={mybluebox},outer arc=2mm,colupper=black, valign=center, halign=left, boxsep = 0pt, title={\textbf{Goal: Interpret neural network encodings}}, halign title=center, segmentation style={solid, line width=1pt}, left=0pt]
        \def\arraystretch{1.0}%

    \includegraphics[width=\columnwidth]{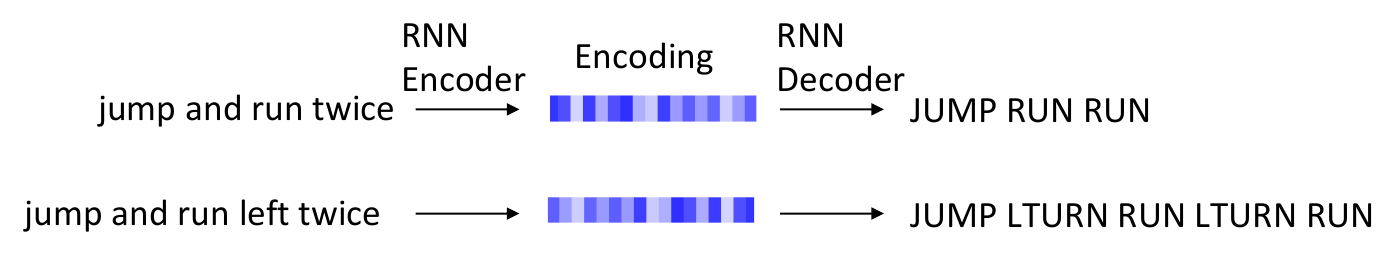}
    \end{tcolorbox}
    
    \hypersetup{linkcolor=white}
    
    \begin{tcolorbox}[width=\columnwidth, colback={myblue}, coltitle=white, toptitle=3pt, bottomtitle=3pt, colframe={mybluebox},outer arc=2mm,colupper=black, valign=center, halign=left, boxsep = 0pt, title={\textbf{Method: Approximate the encodings of a neural network with a more \\ interpretable compositional model}  (\S \color{white}\ref{sec:Role-Learning})}, halign title=center, segmentation style={solid, line width=1pt}, left=3pt]
        \def\arraystretch{1.0}%
    
    \textbf{Step 1: Assign structural roles to words using a learned role assigner.} 
    
    \includegraphics[width=\columnwidth]{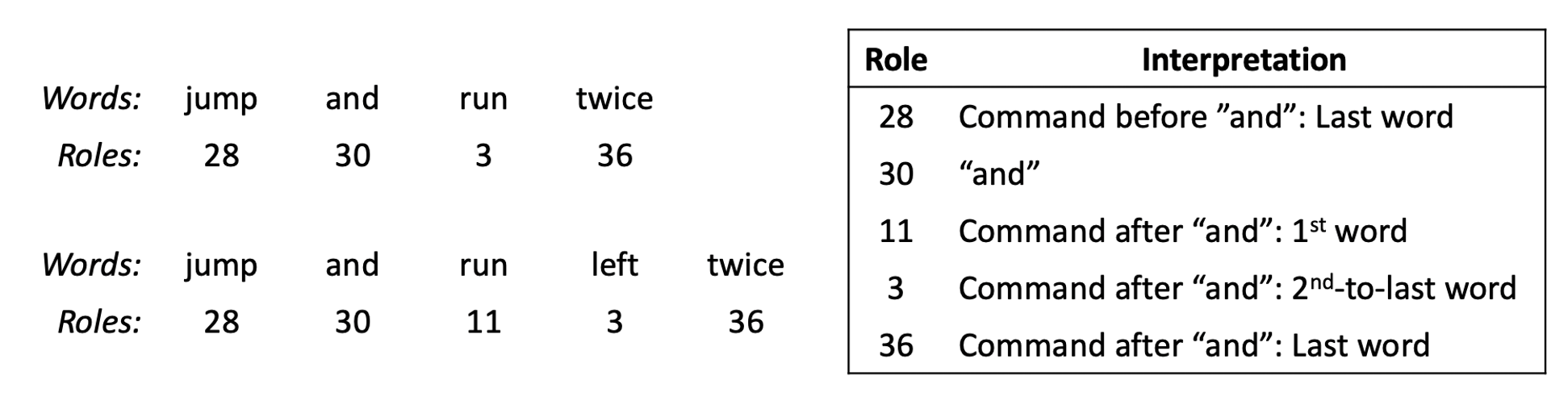}
    
   \textbf{Step 2: Combine word and role vectors using a closed-form equation with learned parameters.} 
    
    \includegraphics[width=\columnwidth]{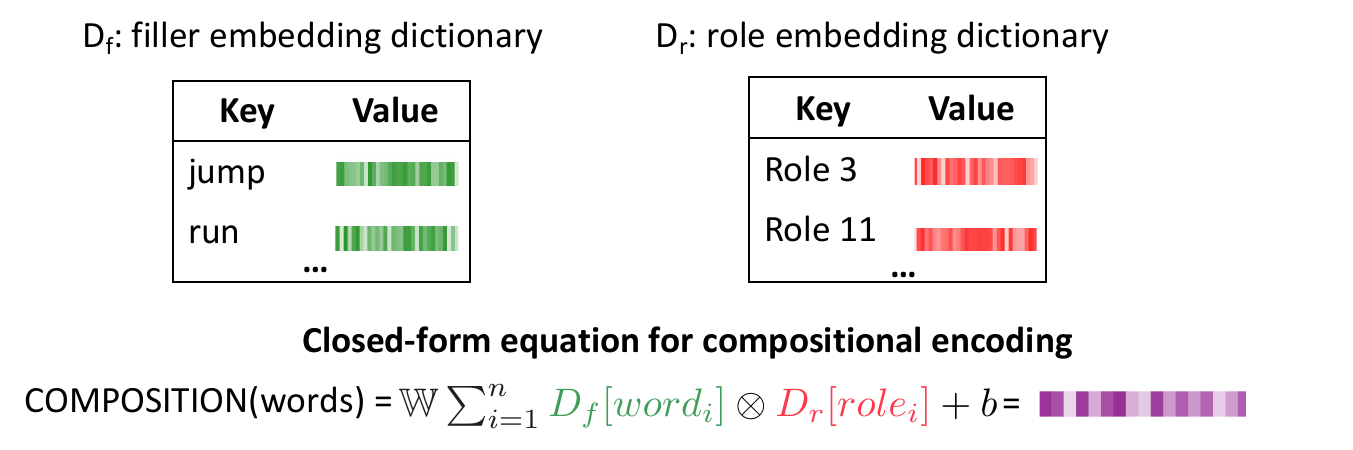}
    
     \end{tcolorbox}
    
    \hypersetup{linkcolor=blue}
    
    \begin{tcolorbox}[width=\columnwidth, colback={myblue}, coltitle=white, toptitle=3pt, bottomtitle=3pt, colframe={mybluebox},outer arc=2mm,colupper=black, valign=center, halign=left, boxsep = 0pt, title={\textbf{Results:}}, halign title=center, segmentation style={solid, line width=1pt}, left=3pt]
        \def\arraystretch{1.0}%
        
    \textbf{The compositional encodings are functionally equivalent to the RNN encodings.} (\S \ref{sec:SCANnetReps})

    \includegraphics[width=\columnwidth]{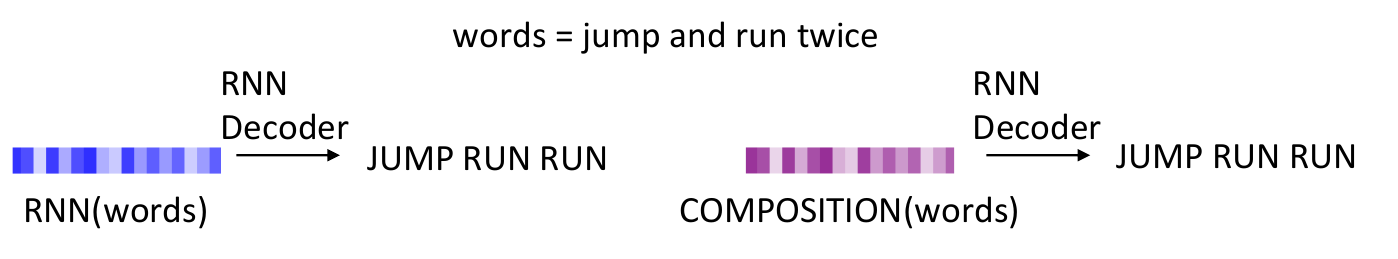}
        
    \textbf{The RNN encodings can be manipulated in symbolic ways to alter the output.} (\S \ref{sec:Surgery})
    
    \includegraphics[width=\columnwidth]{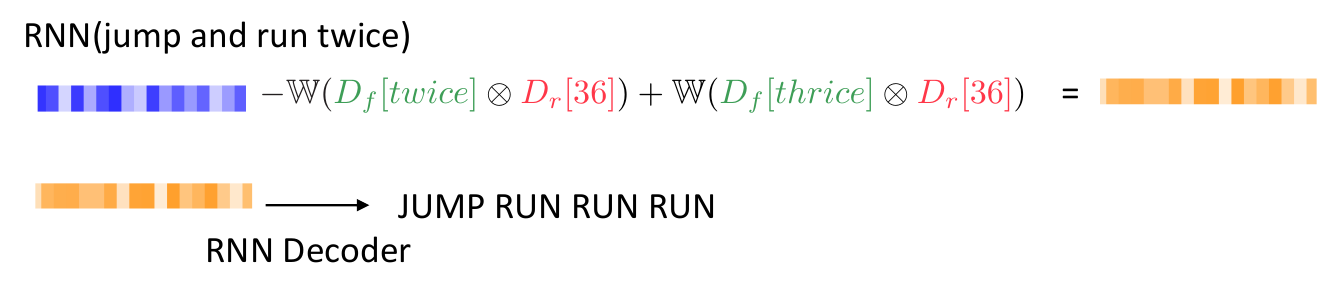}
    
    \end{tcolorbox}
    
    \caption{Summary of our approach.}
    \label{fig:my_label}
\end{figure}


Traditional models of cognition, and language in particular, have relied heavily on symbol structures and symbol manipulation.
However, in the current era, deep learning research has shown that Neural Networks (NNs) can display remarkable degrees of generalization on tasks traditionally viewed as depending on symbolic structure  \citep{googlenmt, mccoy}, albeit with some important limits to their generalization \citep{lake2018generalization}.
Given that standard NNs have no obvious mechanisms for representing symbolic structures, parsing inputs into such structures, nor applying compositional symbol-manipulating rules to them, this success raises the question that we address in this paper:\textbf{\emph{ How do NNs achieve such strong performance on compositional tasks?}} 

Could it be that NNs \textit{do} learn symbolic representations---covertly embedded as vectors in their state spaces?
\citet{mccoy} showed that 
when trained on highly compositional tasks, 
standard NNs learned representations that are functionally equivalent to compositional vector embeddings of symbolic structures (Sec. \ref{sec:TPR}). Processing in these NNs assigns structural representations to inputs and generates outputs that are governed by compositional rules stated over those representations. We refer to the networks we will analyze as \newterm{target NNs}, because we will propose a new type of NN (in Sec.~\ref{sec:Role-Learning})---the \newterm{Role Learner (\RLN)}---which is used to \textit{analyze} the target network.
In contrast with the analysis model of \citet{mccoy}, which relies on a hand-specified hypothesis about the structure underlying the learned representations of the target NN, \RLN\ \textit{automatically} learns a symbolic structure that best approximates the internal representation of the target network. This yields 
two 
advantages. 
First, \RLN\ achieves success at analyzing networks for which the underlying structure is unclear. We show this in Sec.~\ref{sec:SCAN}, where \RLN\ successfully uncovers the symbolic structures learned by a seq2seq RNN trained on the SCAN synthetic semantic parsing task \citep{lake2018generalization}. Second, removing the need for hand-specified structural hypotheses reduces the burden on the analyst, who only needs to provide input sequences and their target NN encodings. Discovering symbolic structure within a model enables us to perform precise alterations to the internal representations in order to produce desired alterations in the output (Sec.~\ref{sec:Surgery}). Then, in Sec.~\ref{sec:NLP}, we turn briefly to partially-compositional tasks in NLP.

The novel contributions of this research are:
\begin{itemize}
  \vspace{-2mm}
  \item ROLE, a NN module that learns to assign symbolic structures to input sequences (Sec.~\ref{sec:Role-Learning}).
  \vspace{-2mm}
  \item Demonstration that RNNs converge to compositional solutions on the synthetic SCAN task (Sec.~\ref{sec:SCAN}).
  \vspace{-2mm}
  \item A precise closed-form expression for the distributed encoding learned by an RNN trained on SCAN, exhibiting its latent symbolic structure (Sec.~\ref{sec:scan-role-interpretation}).
  \vspace{-2mm}
  \item Demonstration of the causal relevance of this symbolic structure by using the equation for its vector encoding to control RNN output through precise alteration of the RNN's internal encoding (Sec.~\ref{sec:Surgery}).
  \vspace{-2mm}
  \item Additional evidence showing that sentence embedding models do not capture compositional structure (Sec. \ref{sec:NLP}).
\end{itemize}

\section{Background Related work} \label{sec:Related}

\subsection{Compositionality}

Certain cognitive tasks consist in computing a function $\varphi$ that is governed by strict rules: e.g., if $\varphi$ is the function mapping a mathematical expression to its value (e.g., mapping `$19 - 2 * 7$' to $5$), then $\varphi$ obeys the rule that $\varphi(x + y) = \mathtt{sum}(\varphi(x), \varphi(y))$ for any expressions $x$ and $y$.
This rule is \newterm{compositional}: the output of a structure (here, $x + y$) is a function of the outputs of the structure's constituents (here, $x$ and $y$). The rule can be stated with full generality once the input is assigned a \newterm{symbolic structure} giving its decomposition into constituents. 
For a \newterm{fully-compositional} task, completely determined by compositional rules, a system that can assign appropriate symbolic structures to inputs and apply appropriate compositional rules to these structures will display full \newterm{systematic generalization}: it will correctly process arbitrary novel combinations of familiar constituents. This is a core capability of symbolic AI systems.

Other tasks, including most natural language tasks such as machine translation, are only partially characterizable by compositional rules: natural language is only partially compositional in nature. For example, if $\varphi$ is the function that assigns meanings to English adjectives, it generally obeys the rule that $\varphi(\mathtt{in\mbox{-}} + x) = \mathtt{not} \hs \varphi(x)$, (e.g., $\varphi(\mathtt{inoffensive}) = \mathtt{not} \hs \varphi(\mathtt{offensive})$), yet there are exceptions: $\varphi(\mathtt{inflammable}) = \varphi(\mathtt{flammable})$. On these ``\newterm{partially-compositional}'' tasks, this strategy of compositional analysis has demonstrated considerable, but limited, generalization capabilities.

\subsection{Analysis of NNs}

Many past works in the rich body of literature about analyzing NNs focus on compositional structure
\citep{hupkes2020compositionality, hupkes2018visualisation, hewitt2019structural, li-etal-2019-compositional} and systematicity \citep{lake2018generalization,goodwin2020probing}. 
Two of the most popular analysis techniques are the behavioral and probing approaches. In the behavioral approach, a model is evaluated  on a set of examples carefully chosen to require competence in particular linguistic phenomena \citep{marvin2018targeted, wang2018glue, dasgupta2019analyzing, poliak2018collecting,linzen2016assessing,mccoy2019right,warstadt2019blimp}. This technique can illuminate behavioral shortcomings but says little about how the internal representations are structured, treating the model as a black box.

In the probing approach, an auxiliary classifier is trained to classify the model's internal representations based on some linguistically-relevant distinction \citep{adi2016fine,giulianelli2018hood,conneau2018cram,conneau2018senteval,belinkov2017evaluating,blevins2018hierarchical,peters2018dissecting,tenney2018what}. In contrast with the behavioral approach, the probing approach tests whether some particular information is present in the model's encodings, but it says little about whether this information is actually used by the model. Indeed, in some cases models fail despite having the necessary information to succeed in their representations, showing that the ability of a classifier to extract that information does not mean that the model is using it \citep{ Voita2020InformationTheoreticPW, ravichander2020probing,vanmassenhove2017investigating}.

We build on \citet{mccoy}, which introduced the analysis task \newterm{DISCOVER (DISsecting COmpositionality in VEctor Representations)}:  take a NN and, to the extent possible, find an explicitly-compositional approximation to its internal distributed representations. DISCOVER allows us to bridge the gap between representation and behavior: It reveals not only what information is encoded in the representation, but also reveals this information in a way that we can manipulate to show that the information is causally implicated in the model's behavior (Section~\ref{sec:Surgery}). Moreover, it provides a much more comprehensive window into the representation than the probing approach does; while probing extracts particular types of information from a representation (e.g., ``does this representation distinguish between active and passive sentences?''), DISCOVER exhaustively decomposes the model's representational space. In this regard, DISCOVER is most closely related to the approaches of \citet{andreas2019measuring}, \citet{chrupala2019correlating}, and \citet{abnar2019blackbox}, who also propose methods for discovering a complete symbolic characterization of a set of vector representations, and \citet{omlin1996extraction} and \citet{weiss2018extracting}, which also seek to extract  more interpretable symbolic models that approximate neural network behavior.
Like \citet{andreas2019measuring} and \citet{chrupala2019correlating}, we seek to find the structure encoded in neural networks, rather than seeking structure directly from the data as is the goal in grammar induction work such as \citet{shen2018ordered} and \citet{bowman2016fast}.

\section{NN embedding of symbol structures} \label{sec:TPR}

\begin{figure}[t]
    \centering
    \includegraphics[scale=.25]{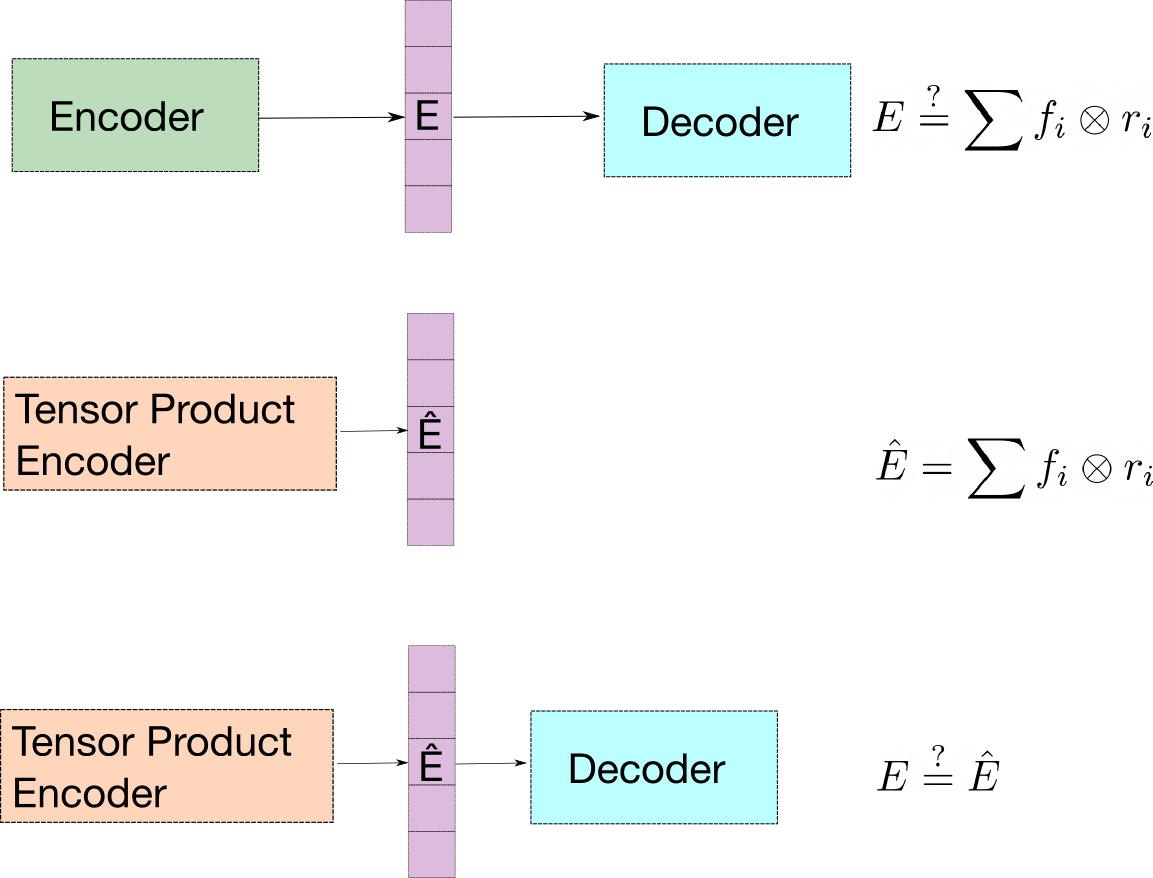}
    \caption{The DISCOVER task and functions. At the top is the target network and question we pose: is the internal embedding a TPR? The middle row is the TPE which follows the provided equation. We train the TPE to minimize the MSE between $\hat{E}$ and $E$. In the bottom row, we evaluate our model by passing the approximations $\hat{E}$ through the decoder and checking the \textit{substitution accuracy} --- the proportion of examples for which the approximated encoding $\hat{E}$ yields the correct output when provided to the decoder .}
    \label{fig:discover}
\end{figure}

\citeauthor{mccoy} showed that, in GRU \citep{cho-etal-2014-learning} encoder-decoder networks performing simple, fully-compositional string manipulations, the medial encoding (between encoder and decoder) could be extremely well approximated, up to an affine transformation, by \newterm{Tensor Product Representations (TPRs)} \citep{Smolensky:1990:TPV:102418.102425}, which are explicitly-compositional vector embeddings of symbolic structures. To represent a string of symbols as a TPR, the symbols in the string $\t{337}$ might be parsed into three constituents $\{ \t{3}\colon\!\r{pos}1, \t{7}\colon\!\r{pos}3, \t{3}\colon\!\r{pos}2 \}$, where $\r{pos}n$ is the role of $n^{\r{th}}$ position from the left edge of the string; other role schemes are also possible, such as roles denoting right-to-left position: $\{ \t{3}\colon\!\r{third\mbox{-}to\mbox{-}last}, \t{3}\colon\!\r{second\mbox{-}to\mbox{-}last}, \t{7}\colon\!\r{last} \}$. The embedding of a constituent $\t{7}\colon\!\r{pos}3$ is $\b{e}(\t{7}\colon\!\r{pos}3) = \b{e}_\rF(\t{7}) \otimes \b{e}_\rR(\r{pos3})$, where $\otimes$ is the tensor product (outer product), $\b{e}_\rR, \b{e}_\rF$ are respectively a vector embedding of the roles and a vector embedding of the \newterm{fillers} of those roles: the digits. The embedding of the whole string is the sum of the embeddings of its constituents. 
In general, for a symbol structure $\t{S}$ with roles $\{ r_k \}$ that are respectively filled by the symbols $\{ \t{f}_k \}$,
$\b{e}_\r{TPR}(\t{S}) = \sum_k \b{e}_\rF(\t{f}_k) \otimes \b{e}_\rR(r_k)$. The DISCOVER task including the TPR equations is depicted in Figure \ref{fig:discover}.

At a high level, these role embeddings serve a similar purpose as positional embeddings in a Transformer \citep{vaswani2017attention}, in that they are vector embeddings of a token's position in a sequence. The roles discussed above---and the positional embeddings used in Transformers---illustrate \textbf{role schemes} based on sequential position; non-sequential role schemes such as positions in a tree are also possible. \citet{mccoy} showed that,for a given seq2seq architecture learning a given string-mapping task, there exists a highly accurate TPR approximation of the medial encoding, given an appropriate  pre-defined role scheme. The main technical contribution of the present paper is the Role Learner (ROLE) model, an RNN that learns its own role scheme to optimize the fit of a TPR approximation to a given set of internal representations in a pre-trained target NN. This makes the DISCOVER framework more general by removing the need for human-generated hypotheses about the role schemes the network might be implementing. Learned role schemes, we will see in Sec.~\ref{sec:SCANnetReps}, can enable good TPR approximation of networks for which human-generated role schemes fail.

\section{The Role Learner (ROLE) Model} \label{sec:Role-Learning} 

\RLN\footnote{Code available at \url{https://github.com/psoulos/role-decomposition}.} produces a vector-space embedding of an input string of $T$ symbols $\t{S} = \t{s}_1 \t{s}_2 \ldots \t{s}_T$ by producing a TPR $\b{T}(\t{S})$ and then passing it through an affine transformation.
\RLN\ is trained to approximate a pre-trained target string-encoder $\c{E}$.
Given a set of $N$ training strings $\{ \t{S}^{(1)}, \ldots, \t{S}^{(N)} \}$, \RLN\ minimizes the total mean-squared error (MSE) between its output $\bW\,\b{T}(\t{S}^{(i)}) + \vb$ and 
$\c{E}(\t{S}^{(i)})$.

\begin{figure}[t]
    \centering
    \includegraphics[scale=.3]{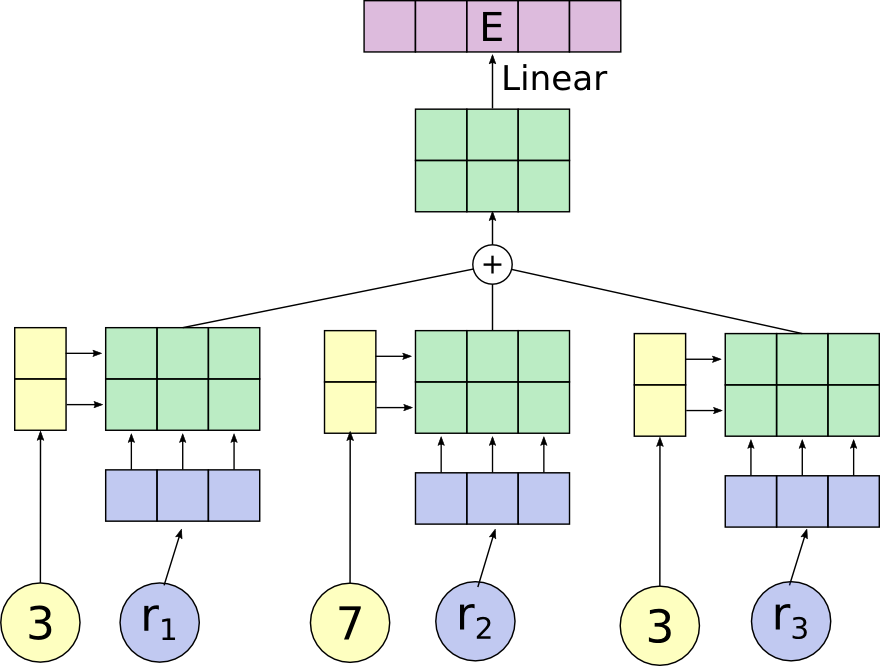}
    \caption{The Tensor Product Encoder architecture. The fillers (yellow circles) and roles (blue circles) are first vectorized with an embedding layer. These two vector embeddings are combined by an outer product to produce the green matrix representing the TPR of the constituent. All of the constituents are summed together to produce the TPR of the sequence, and then a linear transformation is applied to resize the TPR to the target encoder's dimensionality. \RLN\ replaces the role embedding layer and directly produces the blue role vector.}
    \label{fig:tpe-arch}
\end{figure}

\RLN\ is an extension of the Tensor-Product Encoder (TPE) introduced in \citet{mccoy} (as the ``Tensor Product Decomposition Network'') and depicted in Figure \ref{fig:tpe-arch}. 
Crucially, \RLN\ is not \textit{given} role labels for the input symbols, but \textit{learns to compute} them.
More precisely, it learns a dictionary of $n_\rR$ $d_\rR$-dimensional role-embedding vectors, 
$\mR \in \R^{d_\rR \times n_\rR}$, and, for each input symbol 
$\t{s}_t$, computes a soft-attention vector $\va_t$ over these role vectors: 
the role vector assigned to $\t{s}_t$ is then the attention-weighted linear combination of role vectors, $\vr_t = \mR\, \va_t$. \RLN\ simultaneously learns a dictionary of $n_\rF$ $d_\rF$-dimensional symbol-embedding filler vectors $\mF \in \R^{d_\rF \times n_\rF}$, the $\phi^{\r{th}}$ column of which is $\vf_\phi$, the embedding of symbol type $\phi$; $\phi \in 1, \ldots, n_\rF$ where $n_\rF$ is the size of the vocabulary of symbol types.
The TPR generated by \RLN\ is thus $\b{T}(\t{S}) = \sum_{t=1}^T \vf_{\tau(\t{s}_t)} \otimes \vr_t$, where $\tau(\t{s}_t)$ is symbol $\t{s}_t$'s type.
Finally, \RLN\ learns an affine transformation to map this TPR into $\R^d$, where $d$ is the dimension of the representations of the encoder $\c{E}$.

\begin{figure}[t]
    \captionsetup{width=\columnwidth}
    \includegraphics[width=\columnwidth]{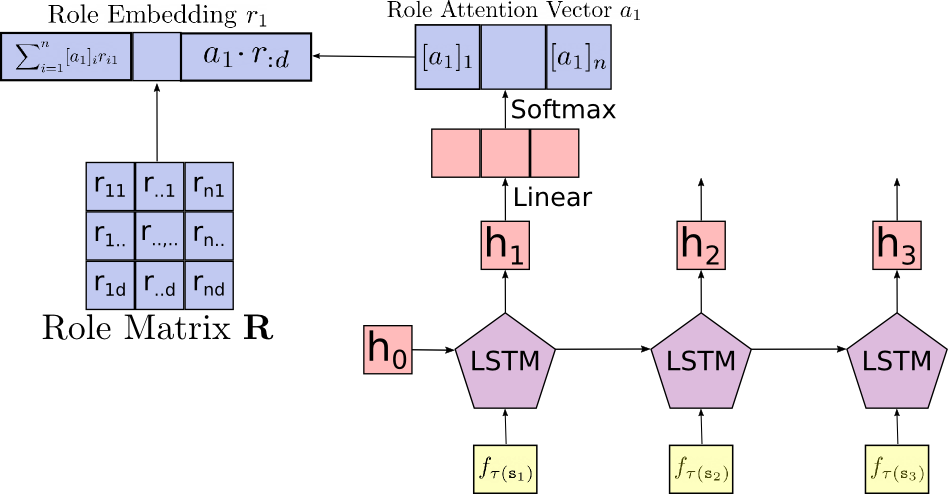}
    \caption{The role learning module. The role attention vector $a_t$ is encouraged to be one-hot through regularization; if $a_t$ were one-hot, the produced role embedding $r_t$ would correspond directly to one of the roles defined in the role matrix $\mR$. The LSTM can be unidirectional or bidirectional.}
    \label{fig:role_learner}
\end{figure}

\RLN\ uses an LSTM \citep{Hochreiter:1997:LSM:1246443.1246450} to compute the role-assigning attention-vectors $\va_t$ from its learned embedding $\mF$ of the input symbols $\t{s}_t$: at each $t$, the hidden state of the LSTM passes through a linear layer and then a softmax to produce $\va_t$ (depicted in Figure \ref{fig:role_learner}). Let the $t^{\r{th}}$ LSTM hidden state be $\vq_t \in \R^H$; let the output-layer weight-matrix have rows $\vk_\rho^\top \in \R^H$ and let the columns of $\mR$ be $\vv_\rho \in R^{d_\rR}$, with $\rho = 1,\dots,n_\rR$. Then $\vr_t = \mR\, \va_t = \sum_{\rho=1}^{n_\rR} \vv_\rho\, \softmax ( \vk_\rho^\top \vq_t )$: the result of query-key attention  \citep[e.g.,][]{vaswani2017attention} with query $\vq_t$ to a fixed external memory containing key-value pairs $\{ ( \vk_\rho, \vv_\rho ) \}_{\rho=1}^{n_\rR}$.

Since a TPR for a discrete symbol structure deploys a discrete set of roles specifying discrete structural positions, ideally a single role would be selected for each $\t{s}_t$: $\va_t$ would be one-hot.
\RLN\ training therefore deploys regularization to bias learning towards one-hot $\va_t$ vectors (based on the regularization proposed in \citet{palangi}, developed for the same purpose). See Appendix \ref{sec:role-regulatization} for the precise regularization terms that we used.

It is essential to note that, while we impose this regularization on \RLN, there is no explicit bias favoring discrete compositional representations in the \textit{target encoder} $\c{E}$: any such structure that \RLN\ finds hidden in the representations learned by $\c{E}$ must result from biases implicit in the vanilla RNN-architecture of $\c{E}$ when applied to its target task.

\section{The SCAN task} \label{sec:SCAN}

Returning to our central question from Sec.~\ref{sec:Intro}, how can neural networks \textit{without} explicit compositional structure perform well on fully-compositional tasks? 
Our hypothesis is that, though these models have no \textit{constraint} forcing them to be compositional, they still have the \textit{ability} to implicitly learn compositional structure.
To test this hypothesis, we apply \RLN\ to a standard RNN-based seq2seq model \citep{sutskever2014sequence} trained on a fully compositional task. Because the RNN has no constraint forcing it to use TPRs, we do not know \textit{a priori} whether there exists any solution that \RLN\ could learn; thus, if \RLN\ does learn anything it will be a significant empirical finding about how these RNNs operate.

We consider the SCAN task \citep{lake2018generalization}, which was designed to test compositional generalization and systematicity. SCAN is a synthetic semantic parsing
task: an input sequence describing an action plan, e.g., \textcolor{violet}{$\t{jump \hs opposite \hs left}$}, is mapped to a sequence of primitive actions, e.g., \textcolor{violet}{$\t{TL \hs TL \hs JUMP}$} (see Sec.~\ref{sec:Surgery} for a complex example). We use \textcolor{violet}{$\t{TL}$} to abbreviate \textcolor{violet}{$\t{TURN\_LEFT}$}, sometimes written \textcolor{violet}{$\t{LTURN}$}; similarly, we use \textcolor{violet}{$\t{TR}$} for \textcolor{violet}{$\t{TURN\_RIGHT}$}. The SCAN mapping is defined by a complete set of compositional rules \citep[Supplementary Fig. 7]{lake2018generalization}. 

\begin{table*}[ht]
\resizebox{\textwidth}{!}{
\begin{tabular}{ c c c c c c c c c} 
 \toprule
  \textbf{Continuous} & \textbf{Snapped} & \textbf{Discrete} & LTR & RTL & Bi & Tree & Wickel & BOW \\ 
  \midrule
  94.83\% & 81.71\% $\pm$ 7.28 & 92.44\% & 6.68\% & 6.96\% & 10.72\% & 4.31\% & 44.00\% & 4.52\% \\
 \bottomrule
\end{tabular}
}
\caption{\label{tab:scan-accuracy}Mean substitution accuracy for learned (bold) and hand-defined role schemes on SCAN across three random initializations. Standard deviation was below 1\% for all schemes except for snapped. Substitution accuracy is measured by feeding \RLN's approximation to the target decoder. (Sec.~\ref{sec:SCANnetReps})}
\end{table*}

\subsection{The compositional structure of SCAN encoder representations} \label{sec:SCANnetReps}

For our target SCAN encoder $\c{E}$, we trained a standard GRU with one hidden layer of dimension 100 for 100,000 steps (batch-size 1) with a dropout of 0.1 on the simple train-test split (hyperparameters determined by a limited search; see Appendix \ref{sec:rnn-scan}).
$\c{E}$ achieves $98.47\%$ (full-string) accuracy on the test set. Thus $\c{E}$ provides what we want: a standard RNN achieving near-perfect accuracy on a non-trivial fully compositional task.

After training, we extract the final hidden embedding from the encoder for each example in the training and test sets. These are the encodings we attempt to approximate as explicitly compositional TPRs. 
We provide \RLN\ with 50 roles to use as it wants (hyperparameters described in Appendix \ref{sec:role-scan}).
We evaluate the substitution accuracy that this learned role scheme provides in three ways. The \newterm{continuous} method tests ROLE in the same way as it was trained, with input symbol $\t{s}_t$ assigned role vector $\vr_t = \mR\, \va_t$. The continuous method does not produce a discrete set of role vectors because the linear layer that generates $a_t$ allows for continuously-valued weights. The remaining two methods test the efficacy of a truly discrete set of role vectors. First, in the \newterm{snapped} method, $\va_t$ is replaced at evaluation time by the one-hot vector $\vm_t$ singling out role $m_t = \argmax(\va_t)$: $\vr_t = \mR\, \vm_t$. This method serves the goal of enforcing the discreteness of roles, but it is expected to decrease performance because it tests ROLE in a different way than it was trained. Our final evaluation method, the \newterm{discrete} method, uses discrete roles without having such a train/test discrepancy by using a two-stage process. In the first stage, the snapped method is used to output one-hot vector roles $\vm_t$ for every symbol in the dataset. In the second stage, we train a TPE which does not learn roles but rather uses the one-hot vector $\vm_t$ as input during training. In this case, \RLN\ acts as an automatic data labeler, assigning a role to every input word.

For comparison, we also train TPEs using a variety of discrete hand-crafted role schemes: left-to-right (LTR), right-to-left (RTL), bidirectional (Bi), tree position, neighbor-based Wickelrole (Wickel), and bag-of-words (BOW) (descriptions of these role schemes are in Appendix \ref{sec:role-schemes}). 

The mean substitution accuracy from these different methods is shown in Table \ref{tab:scan-accuracy}. 
All of the predefined role schemes provide poor approximations, none surpassing $44.00\%$ accuracy. 
The role scheme learned by \RLN\ does significantly better than any of the predefined role schemes: when tested with the basic, continuous role-attention method, the accuracy is $94.83\%$. 

The success of \RLN\ 
tells us two things. First, it shows that the target model's compositional behavior relies on compositional internal representations: it was by no means guaranteed to be the case that \RLN\ would be successful here, so the fact that it is successful tells us that the encoder has learned compositional representations. Second, it adds further validation to the efficacy of \RLN, because it shows that it can be a useful analysis tool in cases of significantly greater complexity than the simple string manipulation tasks studied in \citet{mccoy}. 
In fact, it allows us to \textit{\textbf{write in closed form the embedding}} $\b{e}(\t{S})$ of an input $\t{S} = \t{s}_1\ldots\t{s}_T$ that is learned by the SCAN encoder, to an excellent degree of approximation (as measured by substitution accuracy):
$\b{e}(\t{S}) = \bW \sum_{t=1}^T \vf_{\tau(\t{s}_t)} \otimes \vr_{\rho(\t{s}_t)} + \vb$, where $\tau(\t{s}_t)$ is symbol $\t{s}_t$'s type, $\rho(\t{s}_t)$ is the role assigned to $\t{s}_t$ by the algorithm discussed next,
and the matrices $\bW$, $\mF = [\vf_1 \ldots \vf_{n_\rF} ]$, and $\mR = [\vr_1 \ldots \vr_{n_\rR}]$ and bias vector $\vb$ are learned by \RLN.
Note that this expression is bilinear, even though the GRU encoder that generates it includes nonlinearities.

\subsection{Interpreting the learned role scheme}\label{sec:scan-role-interpretation}

By analyzing the roles assigned by ROLE to the sequences in the SCAN training set, we created a symbolic algorithm for predicting which role will be assigned to each filler. This section covers the primary factors of the algorithm, while the entire algorithm is described in Appendix \ref{sec:Alg} and discussed at additional length in Appendix \ref{sec:AlgDisc}. Though the algorithm was created based only on sequences in the SCAN training set, it is equally successful at predicting which roles will be assigned to test sequences, exactly matching \RLN's predicted roles for 98.7\% of sequences.

The algorithm illuminates how the filler-role scheme encodes information relevant to the task. First, one of the initial facts that the decoder must determine is whether the sequence is a single command, a pair of subcommands connected by \textcolor{violet}{$\t{and}$}, or a pair of subcommands connected by \textcolor{violet}{$\t{after}$}; such a determination is crucial for knowing the basic structure of the output (how many actions to perform and in what order). We have found that role 30 is used for, and only for, the filler \textcolor{violet}{$\t{and}$}, while role 17 is used in and only in sequences containing \textcolor{violet}{$\t{after}$} (usually with \textcolor{violet}{$\t{after}$} as the filler bound to role 17). Thus, the decoder can use these roles to tell which basic structure is in play: if role 30 is present, it is an \textcolor{violet}{$\t{and}$} sequence; if role 17 is present, it is an \textcolor{violet}{$\t{after}$} sequence; otherwise it is a single command.

Once the decoder has established the basic syntactic structure of the output, it must then fill in the particular actions. This can be accomplished using the remaining roles, which mainly encode absolute position within a subcommand. For example, the last word of a subcommand before \textcolor{violet}{$\t{after}$} (e.g., \textcolor{violet}{$\t{jump \hs \textbf{left} \hs after \hs walk \hs twice}$}) is always assigned role 8, while the last word of a subcommand after \textcolor{violet}{$\t{after}$} (e.g., \textcolor{violet}{$\t{jump \hs left \hs after \hs walk \hs \textbf{twice}}$}) is always assigned role 46. Therefore, once the decoder knows (based on the presence of role 17) that it is dealing with an \textcolor{violet}{$\t{after}$} sequence, it can check for the fillers bound to roles 8 and 46 to begin to figure out what the two subcommands surrounding \textcolor{violet}{$\t{after}$} look like. The identity of the last word in a subcommand is informative because that is where a cardinality (i.e., \textcolor{violet}{$\t{twice}$} or \textcolor{violet}{$\t{thrice}$}) appears if there is one. Thus, by checking what filler is at the end of a subcommand, the model can determine whether there is a cardinality present and, if so, which one.

ROLE itself does not provide an interpretation for the symbolic structure it generates, but
we have shown that this structure can be successfully interpreted by humans. By contrast, it is very difficult to interpret the continuous neuron values of RNN representations; even the rare successful cases of doing so, such as \citet{lakretz2019emergence} and \citet{mu2020compositional}, only interpret a few isolated units, while we were able to exhaustively explain the entire symbolic structure discovered by ROLE.

\begin{figure*}[ht]
    \begin{center}
    \begin{subfigure}{0.5\textwidth}
    \fontsize{7}{8}\selectfont
    $   
    \t{run}\colon\!11 \hs \t{left}\colon\!36 \hs \t{twice}\colon\!8 \hs \t{after}\colon\!43 \hs \t{jump}\colon\!10 \hs \t{opposite}\colon\!17 \hs \t{right}\colon\!4 \hs \t{thrice}\colon\!46 \rightarrow \t{\hs\hs\hs\hs\hs\hs\hs\hs}\t{TR \hs TR \hs JUMP \hs TR \hs TR \hs JUMP \hs TR \hs TR \hs JUMP \hs TL \hs RUN \hs TL \hs RUN} \leavevmode\newline -\t{run}\colon\!11 \hs +\t{look}\colon\!11 \rightarrow \leavevmode\newline \t{\hs\hs\hs\hs\hs\hs\hs\hs}\t{TR \hs TR \hs JUMP \hs TR \hs TR \hs JUMP \hs TR \hs TR \hs JUMP \hs TL \hs \hi{LOOK} \hs TL \hs \hi{LOOK}} \leavevmode\newline
    -\t{jump}\colon\!10 \hs +\t{walk}\colon\!10 \rightarrow\leavevmode\newline
    \t{\hs\hs\hs\hs\hs\hs\hs\hs}\t{TR \hs TR \hs \hi{WALK} \hs TR \hs TR \hs \hi{WALK} \hs TR \hs TR \hs \hi{WALK} \hs TL \hs LOOK \hs TL \hs LOOK} \leavevmode\newline
    -\t{left}\colon\!36 \hs +\t{right}\colon\!36 \rightarrow\leavevmode\newline
    \t{\hs\hs\hs\hs\hs\hs\hs\hs}\t{TR \hs TR \hs WALK \hs TR \hs TR \hs WALK \hs TR \hs TR \hs WALK \hs \hi{TR} \hs LOOK \hs \hi{TR} \hs LOOK} \leavevmode\newline
    -\t{twice}\colon\!8 \hs +\t{thrice}\colon\!8 \rightarrow\leavevmode\newline
    \t{\hs\hs\hs\hs\hs\hs\hs\hs}\t{TR \hs TR \hs WALK \hs TR \hs TR \hs WALK \hs TR \hs TR \hs WALK \hs TR \hs LOOK \hs } \leavevmode\newline
    \t{\hs\hs\hs\hs\hs\hs\hs\hs}\t{TR \hs LOOK \hs \hi{TR \hs LOOK}} \leavevmode\newline
    -\t{opposite}\colon\!17 \hs +\t{around}\colon\!17 \rightarrow\leavevmode\newline
    \t{\hs\hs\hs\hs\hs\hs\hs\hs}\hi{\t{TR \hs WALK \hs TR \hs WALK \hs TR \hs WALK \hs TR \hs WALK \hs TR \hs WALK \hs TR \hs WALK}} \leavevmode\newline
    \t{\hs\hs\hs\hs\hs\hs\hs\hs}\hi{\t{TR \hs WALK \hs TR \hs WALK \hs TR \hs WALK \hs TR \hs WALK \hs \hi{TR \hs WALK \hs TR \hs WALK} \hs }} \leavevmode\newline
    \t{\hs\hs\hs\hs\hs\hs\hs\hs}\t{TR \hs LOOK \hs TR \hs LOOK \hs TR \hs LOOK} \leavevmode\newline
$
        \label{fig:role-distribution}
    \end{subfigure}
    \hfill
    \begin{subfigure}{0.28\textwidth}
        \hspace{-40pt}
        \includegraphics[width=\textwidth]{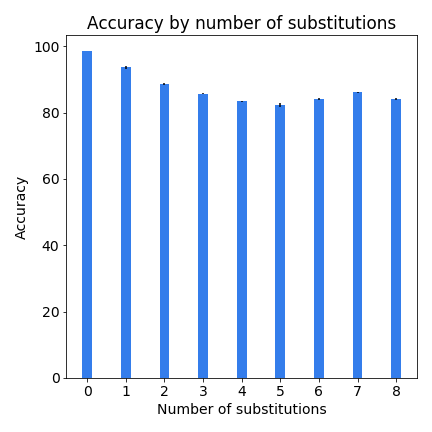}
        \label{fig:surgery}
    \end{subfigure}
    \caption{Left: Example of successive constituent surgeries. The roles assigned to the input symbols are indicated in the first line (e.g., \textcolor{violet}{$\t{run}$} was assigned role $11$). Altered output symbols are in blue. The model produces the correct outputs for all cases shown here. Right: Mean constituent-surgery accuracy across three runs. Standard deviation is below 1\% for each number of substitutions. (Sec.~\ref{sec:Surgery})}
    \label{fig:Figs12}
    \end{center}
\end{figure*}

\subsection{Precision constituent-surgery on internal representations produces desired outputs} \label{sec:Surgery}

The substitution-accuracy results above show that if the \textit{entire} learned representation is replaced by \RLN's approximation, the output remains correct. But do the \textit{individual word embeddings} in this TPR have the appropriate causal consequences when processed by the decoder?

To address this causal question \citep{pearl2000causality}, we actively intervene on the constituent structure of the internal representations by replacing one constituent with another syntactically equivalent
one,\footnote{We extract syntactic categories from the SCAN grammar \citep[Supplementary Fig. 6]{lake2018generalization} by saying that two words belong to the same category if every occurrence of one could be grammatically replaced by the other. We do not replace occurrences of \textcolor{violet}{$\t{and}$} and \textcolor{violet}{$\t{after}$} since the presence of either of these words causes substantial changes in the roles assigned within the sequence (Appendix \ref{sec:Alg}).} and see whether this produces the expected change in the output of the decoder.
We take the encoding generated by the RNN encoder $\c{E}$ for an input such as \textcolor{violet}{$\t{jump \hs opposite \hs left}$}, subtract the vector embedding of the \textcolor{violet}{$\t{opposite}$} constituent, add the embedding of the \textcolor{violet}{$\t{around}$} constituent, and see whether this causes the output to change from the correct output for \textcolor{violet}{$\t{jump \hs opposite \hs left}$ ($\t{TL \hs TL \hs JUMP)}$} to the correct output for \textcolor{violet}{$\t{jump \hs around \hs left}$ ($\t{TL \hs JUMP \hs TL \hs JUMP \hs TL \hs JUMP \hs TL \hs JUMP}$)}. The roles in these constituents are determined by the algorithm of Appendix \ref{sec:Alg}. If changing a word leads other roles in the sequence to change (according to the algorithm), we update the encoding with those new roles as well. Such surgery can be viewed as a more general extension of the analogy approach used by \citet{mikolov2013linguistic} for analysis of word embeddings.
An example of applying a sequence of five such constituent surgeries to a sequence is shown in Figure~\ref{fig:Figs12} (left). Even long sequences of such replacements produce the expected change in the decoder's output with high accuracy (Figure~\ref{fig:Figs12}, right), indicating that the compositional structure discovered by ROLE does play a central causal role in the model's behavior.

\section{Partially-compositional NLP tasks} \label{sec:NLP}
The previous sections explored fully-compositional tasks where there is a strong signal for compositionality. In this section, we explore whether the representations of NNs trained on tasks that are only partially-compositional also capture compositional structure. Partially-compositional tasks are especially challenging to model because a fully-compositional model may enforce compositionality too strictly to handle the non-compositional aspects of the task, while a model without a compositional bias may not learn any sort of compositionality from the weak cues in the training set.

We test four sentence encoding models for compositionality: InferSent \citep{conneau2017supervised}, Skip-thought \citep{kiros2015skip}, Stanford Sentiment Model (SST) \citep{socher2013recursive}, and SPINN \citep{bowman2016fast}. For each of these models, we extract the encodings for the SNLI premise sentences \citep{bowman2015large}. We use the extracted embeddings to train \RLN\ with 50 roles available (additional training information provided in Appendix \ref{sec:role-sentences}).

\begin{table*}[h!]
    \centering
   \resizebox{\textwidth}{!}{
		\begin{tabular} {lcccccccc}
			\toprule
		 & \textbf{Continuous} & \textbf{Snapped} & \textbf{Discrete} & LTR & RTL & Bi & Tree & BOW \\ \midrule
		InferSent & \textbf{4.05e-4} & 4.15e-4 & 5.76e-4 & 8.21e-4 & 9.70e-4 & 9.16e-4 & 7.78e-4 & 4.34e-4 \\
		Skip-thought & 9.30e-5 & 9.32e-5 & 9.85e-5 & 9.91e-5 & 1.78e-3 & 3.95e-4 & 9.64e-5 & \textbf{8.87e-5} \\
		SST & \textbf{5.58e-3} & 6.72e-3 & 6.48e-3 & 8.35e-3 & 9.29e-3 & 8.55e-3 & 5.99e-3 & 9.38e-3 \\
		SPINN & \textbf{.139} & .151 & .147 & .184 & .189 & .181 & .178 & .176 \\ \bottomrule
		\end{tabular}
		}
    \caption{\label{tab:sent-emb-mse}MSE loss for learned (bold) and hand-crafted role schemes on sentence embedding models. (Sec.~\ref{sec:NLP})}

\end{table*}

As a baseline, we also train TPEs that use pre-defined role schemes (hyperparameters described in Appendix \ref{sec:tpe-sentences}). For all of the sentence embedding models except Skip-thought, \RLN\ with continuous attention provides the lowest mean squared error at approximating the encoding (Table~\ref{tab:sent-emb-mse}). The BOW (bag-of-words) role scheme represents a TPE that uses a degenerate `compositional' structure which assigns the same role to every filler; for each of the sentence embedding models tested except for SST, performance is  within the same order of magnitude as structure-free BOW. \citet{parikh2016decomposable} found that a bag-of-words model scores extremely well on Natural Language Inference despite having no knowledge of word order, showing that structure is not necessary to perform well on the sorts of tasks commonly used to train sentence encoders. Although not definitive, the \RLN\ results provide no evidence that these models' sentence embeddings possess compositional structure.

In future work, it would be interesting to perform a similar analysis on Transformer architectures \citep{vaswani2017attention}. Such models have displayed impressive syntactic generalization \cite{hu2020systematic} and few-shot learning of compositional tasks \cite{brown2020language}, both of which suggest that they learn substantial degrees of compositional structure; thus, ROLE may be more likely to discover meaningful structure in Transformers than in the sentence-embedding models in Table \ref{tab:sent-emb-mse}. Further work has found impressive degrees of syntactic structure in Transformer encodings \cite{hewitt2019structural}, suggesting that there may well be compositional structure for ROLE to pick up on. The main difficulty in applying ROLE to Transformers---and the reason we did not include Transformers in our study---is that the sentence representation used by a Transformer is typically viewed as a variable-sized collection of vectors, whereas ROLE requires single-vector representations; this discrepancy must be overcome if ROLE is to be applied to Transformers. 

One past work \cite{jawahar-etal-2019-bert} has applied ROLE's precursor (the TPDN of \citet{mccoy}) to Transformer representations by choosing the [CLS] token of BERT \cite{devlin2019bert} as the single-vector sentence encoding to decompose. \citeauthor{jawahar-etal-2019-bert} found that these encodings were approximated better by human-specified tree-position roles than by other human-specified candidates (e.g., left-to-right and right-to-left roles). By removing the constraint of requiring human-designed role schemes, ROLE may be able to discover other role schemes that approximate BERT's encodings even more closely.

\section{Conclusion} \label{sec:Conclusion}

We have introduced \RLN, a neural network that learns to approximate the representations of an existing target neural network $\c{E}$ using an explicit symbolic structure. \RLN\ successfully discovers symbolic structure in a standard RNN trained on the fully-compositional SCAN semantic parsing task, even though the RNN has no such structure explicitly present in its architecture. This yields a closed-form equation for the RNN's encoding of any input string. 
When applied to sentence embedding models trained on partially-compositional tasks, \RLN\ performs better than hand-specified hypothesized structures but still provides little evidence that the sentence encodings represent compositional structure. 

While this work has shown that NNs can converge to TPRs to solve compositional tasks, it is still unknown how the weights in the NN actually convert the raw input into a TPR. To investigate this process, in future work we plan to apply our technique to representations of partial sequences. For instance, when the complete input is \textcolor{violet}{$\t{jump \hs right \hs twice}$}, the target RNN must first represent \textcolor{violet}{$\t{jump \hs right}$} as a well-formed TPR at the point when only those two words have been encountered. The representation then needs to be updated when the next word, \textcolor{violet}{$\t{twice}$}, is encountered. By studying the nature of that update, we can gain insight into how the target model builds up a TPR from the input elements.

Uncovering the latent symbolic structure of NN representations learned for fully-compositional tasks is a significant step towards explaining how NNs achieve the level of compositional generalization that they do. In addition, by illuminating shortcomings in the representations learned for standard tasks that are not fully-compositional, ROLE can help suggest types of inductive bias for improving models' generalization with standard, partially-compositional datasets.

\section*{Acknowledgments}

This material is based upon work supported by the National Science Foundation Graduate Research Fellowship Program under Grant No. 1746891, and work partially supported by NSF grant BCS1344269. Any opinions, findings, and conclusions or recommendations expressed in this material are those of the authors and do not necessarily reflect the views of the National Science Foundation. 

For helpful comments we are grateful to the members of the Johns Hopkins Neurosymbolic Computation group and the Microsoft Research AI Deep Learning Group. Any errors remain our own.

\bibliographystyle{blackbox}
\bibliography{blackbox}

\begin{thebibliography}{55}
\expandafter\ifx\csname natexlab\endcsname\relax\def\natexlab#1{#1}\fi

\bibitem[{Abnar et~al.(2019)Abnar, Beinborn, Choenni, and
  Zuidema}]{abnar2019blackbox}
Samira Abnar, Lisa Beinborn, Rochelle Choenni, and Willem Zuidema. 2019.
\newblock \href {https://doi.org/10.18653/v1/W19-4820} {Blackbox meets
  blackbox: Representational similarity {\&} stability analysis of neural
  language models and brains}.
\newblock In \emph{Proceedings of the 2019 ACL Workshop BlackboxNLP: Analyzing
  and Interpreting Neural Networks for NLP}, pages 191--203, Florence, Italy.
  Association for Computational Linguistics.

\bibitem[{Adi et~al.(2017)Adi, Kermany, Belinkov, Lavi, and
  Goldberg}]{adi2016fine}
Yossi Adi, Einat Kermany, Yonatan Belinkov, Ofer Lavi, and Yoav Goldberg. 2017.
\newblock \href {https://openreview.net/pdf?id=BJh6Ztuxl} {Fine-grained
  analysis of sentence embeddings using auxiliary prediction tasks}.
\newblock In \emph{International Conference on Learning Representations}.

\bibitem[{Andreas(2019)}]{andreas2019measuring}
Jacob Andreas. 2019.
\newblock \href {https://openreview.net/forum?id=HJz05o0qK7} {Measuring
  compositionality in representation learning}.
\newblock In \emph{International Conference on Learning Representations}.

\bibitem[{Belinkov et~al.(2017)Belinkov, M{\`a}rquez, Sajjad, Durrani, Dalvi,
  and Glass}]{belinkov2017evaluating}
Yonatan Belinkov, Llu{\'\i}s M{\`a}rquez, Hassan Sajjad, Nadir Durrani, Fahim
  Dalvi, and James Glass. 2017.
\newblock \href {https://www.aclweb.org/anthology/I17-1001} {Evaluating layers
  of representation in neural machine translation on part-of-speech and
  semantic tagging tasks}.
\newblock In \emph{Proceedings of the Eighth International Joint Conference on
  Natural Language Processing (Volume 1: Long Papers)}, pages 1--10, Taipei,
  Taiwan. Asian Federation of Natural Language Processing.

\bibitem[{Blevins et~al.(2018)Blevins, Levy, and
  Zettlemoyer}]{blevins2018hierarchical}
Terra Blevins, Omer Levy, and Luke Zettlemoyer. 2018.
\newblock \href {https://doi.org/10.18653/v1/P18-2003} {Deep {RNN}s encode soft
  hierarchical syntax}.
\newblock In \emph{Proceedings of the 56th Annual Meeting of the Association
  for Computational Linguistics (Volume 2: Short Papers)}, pages 14--19,
  Melbourne, Australia. Association for Computational Linguistics.

\bibitem[{Bowman et~al.(2015)Bowman, Angeli, Potts, and
  Manning}]{bowman2015large}
Samuel~R. Bowman, Gabor Angeli, Christopher Potts, and Christopher~D. Manning.
  2015.
\newblock \href {http://aclweb.org/anthology/D15-1075} {A large annotated
  corpus for learning natural language inference}.
\newblock In \emph{{Proceedings of the 2015 Conference on Empirical Methods in
  Natural Language Processing}}, pages 632--642, Lisbon, Portugal. Association
  for Computational Linguistics.

\bibitem[{Bowman et~al.(2016)Bowman, Gauthier, Rastogi, Gupta, Manning, and
  Potts}]{bowman2016fast}
Samuel~R. Bowman, Jon Gauthier, Abhinav Rastogi, Raghav Gupta, Christopher~D.
  Manning, and Christopher Potts. 2016.
\newblock \href {http://aclweb.org/anthology/P16-1139} {A fast unified model
  for parsing and sentence understanding}.
\newblock In \emph{Proceedings of the 54th Annual Meeting of the Association
  for Computational Linguistics (Volume 1: Long Papers)}, pages 1466--1477.
  Association for Computational Linguistics.

\bibitem[{Brown et~al.(2020)Brown, Mann, Ryder, Subbiah, Kaplan, Dhariwal,
  Neelakantan, Shyam, Sastry, Askell et~al.}]{brown2020language}
Tom~B Brown, Benjamin Mann, Nick Ryder, Melanie Subbiah, Jared Kaplan, Prafulla
  Dhariwal, Arvind Neelakantan, Pranav Shyam, Girish Sastry, Amanda Askell,
  et~al. 2020.
\newblock \href {https://arxiv.org/abs/2005.14165} {Language models are
  few-shot learners}.
\newblock \emph{arXiv preprint arXiv:2005.14165}.

\bibitem[{Cho et~al.(2014)Cho, van Merri{\"e}nboer, Gulcehre, Bahdanau,
  Bougares, Schwenk, and Bengio}]{cho-etal-2014-learning}
Kyunghyun Cho, Bart van Merri{\"e}nboer, Caglar Gulcehre, Dzmitry Bahdanau,
  Fethi Bougares, Holger Schwenk, and Yoshua Bengio. 2014.
\newblock \href {https://doi.org/10.3115/v1/D14-1179} {Learning phrase
  representations using {RNN} encoder{--}decoder for statistical machine
  translation}.
\newblock In \emph{Proceedings of the 2014 Conference on Empirical Methods in
  Natural Language Processing ({EMNLP})}, pages 1724--1734, Doha, Qatar.
  Association for Computational Linguistics.

\bibitem[{Chrupa{\l}a and Alishahi(2019)}]{chrupala2019correlating}
Grzegorz Chrupa{\l}a and Afra Alishahi. 2019.
\newblock \href {https://doi.org/10.18653/v1/P19-1283} {Correlating neural and
  symbolic representations of language}.
\newblock In \emph{Proceedings of the 57th Annual Meeting of the Association
  for Computational Linguistics}, pages 2952--2962, Florence, Italy.
  Association for Computational Linguistics.

\bibitem[{Conneau and Kiela(2018)}]{conneau2018senteval}
Alexis Conneau and Douwe Kiela. 2018.
\newblock \href {https://www.aclweb.org/anthology/L18-1269} {{S}ent{E}val: An
  evaluation toolkit for universal sentence representations}.
\newblock In \emph{Proceedings of the Eleventh International Conference on
  Language Resources and Evaluation ({LREC} 2018)}, Miyazaki, Japan. European
  Language Resources Association (ELRA).

\bibitem[{Conneau et~al.(2017)Conneau, Kiela, Schwenk, Barrault, and
  Bordes}]{conneau2017supervised}
Alexis Conneau, Douwe Kiela, Holger Schwenk, Lo{\"i}c Barrault, and Antoine
  Bordes. 2017.
\newblock \href {http://aclweb.org/anthology/D17-1070} {Supervised learning of
  universal sentence representations from natural language inference data}.
\newblock In \emph{Proceedings of the 2017 Conference on Empirical Methods in
  Natural Language Processing}, pages 670--680. Association for Computational
  Linguistics.

\bibitem[{Conneau et~al.(2018)Conneau, Kruszewski, Lample, Barrault, and
  Baroni}]{conneau2018cram}
Alexis Conneau, German Kruszewski, Guillaume Lample, Lo{\"\i}c Barrault, and
  Marco Baroni. 2018.
\newblock \href {https://doi.org/10.18653/v1/P18-1198} {What you can cram into
  a single {\$}{\&}!{\#}* vector: Probing sentence embeddings for linguistic
  properties}.
\newblock In \emph{Proceedings of the 56th Annual Meeting of the Association
  for Computational Linguistics (Volume 1: Long Papers)}, pages 2126--2136,
  Melbourne, Australia. Association for Computational Linguistics.

\bibitem[{Dasgupta et~al.(2019)Dasgupta, Guo, Gershman, and
  Goodman}]{dasgupta2019analyzing}
Ishita Dasgupta, Demi Guo, Samuel~J Gershman, and Noah~D Goodman. 2019.
\newblock \href {https://arxiv.org/abs/1909.05885} {Analyzing machine-learned
  representations: A natural language case study}.
\newblock \emph{arXiv preprint arXiv:1909.05885}.

\bibitem[{Devlin et~al.(2019)Devlin, Chang, Lee, and
  Toutanova}]{devlin2019bert}
Jacob Devlin, Ming-Wei Chang, Kenton Lee, and Kristina Toutanova. 2019.
\newblock \href {https://doi.org/10.18653/v1/N19-1423} {{BERT}: Pre-training of
  deep bidirectional transformers for language understanding}.
\newblock In \emph{Proceedings of the 2019 Conference of the North {A}merican
  Chapter of the Association for Computational Linguistics: Human Language
  Technologies, Volume 1 (Long and Short Papers)}, pages 4171--4186,
  Minneapolis, Minnesota. Association for Computational Linguistics.

\bibitem[{Giulianelli et~al.(2018)Giulianelli, Harding, Mohnert, Hupkes, and
  Zuidema}]{giulianelli2018hood}
Mario Giulianelli, Jack Harding, Florian Mohnert, Dieuwke Hupkes, and Willem
  Zuidema. 2018.
\newblock \href {https://doi.org/10.18653/v1/W18-5426} {Under the hood: Using
  diagnostic classifiers to investigate and improve how language models track
  agreement information}.
\newblock In \emph{Proceedings of the 2018 {EMNLP} Workshop {B}lackbox{NLP}:
  Analyzing and Interpreting Neural Networks for {NLP}}, pages 240--248,
  Brussels, Belgium. Association for Computational Linguistics.

\bibitem[{Goodwin et~al.(2020)Goodwin, Sinha, and
  O{'}Donnell}]{goodwin2020probing}
Emily Goodwin, Koustuv Sinha, and Timothy~J. O{'}Donnell. 2020.
\newblock \href {https://doi.org/10.18653/v1/2020.acl-main.177} {Probing
  linguistic systematicity}.
\newblock In \emph{Proceedings of the 58th Annual Meeting of the Association
  for Computational Linguistics}, pages 1958--1969, Online. Association for
  Computational Linguistics.

\bibitem[{Hewitt and Manning(2019)}]{hewitt2019structural}
John Hewitt and Christopher~D Manning. 2019.
\newblock \href {https://www.aclweb.org/anthology/N19-1419/} {A structural
  probe for finding syntax in word representations}.
\newblock In \emph{Proceedings of the 2019 Conference of the North American
  Chapter of the Association for Computational Linguistics: Human Language
  Technologies, Volume 1 (Long and Short Papers)}, pages 4129--4138.

\bibitem[{Hochreiter and
  Schmidhuber(1997)}]{Hochreiter:1997:LSM:1246443.1246450}
Sepp Hochreiter and J\"{u}rgen Schmidhuber. 1997.
\newblock \href {https://doi.org/10.1162/neco.1997.9.8.1735} {Long short-term
  memory}.
\newblock \emph{Neural Computation}, 9(8):1735--1780.

\bibitem[{Hu et~al.(2020)Hu, Gauthier, Qian, Wilcox, and
  Levy}]{hu2020systematic}
Jennifer Hu, Jon Gauthier, Peng Qian, Ethan Wilcox, and Roger~P Levy. 2020.
\newblock \href {https://www.aclweb.org/anthology/2020.acl-main.158.pdf} {A
  systematic assessment of syntactic generalization in neural language models}.
\newblock In \emph{Proceedings of the 58th Annual Meeting of the Association
  for Computational Linguistics}, Seattle, Washington. Association for
  Computational Linguistics.

\bibitem[{Hupkes et~al.(2020)Hupkes, Dankers, Mul, and
  Bruni}]{hupkes2020compositionality}
Dieuwke Hupkes, Verna Dankers, Mathijs Mul, and Elia Bruni. 2020.
\newblock Compositionality decomposed: How do neural networks generalise?
\newblock \emph{Journal of Artificial Intelligence Research}, 67:757--795.

\bibitem[{Hupkes et~al.(2018)Hupkes, Veldhoen, and
  Zuidema}]{hupkes2018visualisation}
Dieuwke Hupkes, Sara Veldhoen, and Willem Zuidema. 2018.
\newblock Visualisation and `diagnostic classifiers' reveal how recurrent and
  recursive neural networks process hierarchical structure.
\newblock \emph{Journal of Artificial Intelligence Research}, 61:907--926.

\bibitem[{Jawahar et~al.(2019)Jawahar, Sagot, and
  Seddah}]{jawahar-etal-2019-bert}
Ganesh Jawahar, Beno{\^\i}t Sagot, and Djam{\'e} Seddah. 2019.
\newblock \href {https://doi.org/10.18653/v1/P19-1356} {What does {BERT} learn
  about the structure of language?}
\newblock In \emph{Proceedings of the 57th Annual Meeting of the Association
  for Computational Linguistics}, pages 3651--3657, Florence, Italy.
  Association for Computational Linguistics.

\bibitem[{Kingma and Ba(2015)}]{kingma2015adam}
Diederik Kingma and Jimmy Ba. 2015.
\newblock \href {https://arxiv.org/pdf/1412.6980.pdf} {Adam: A method for
  stochastic optimization}.
\newblock In \emph{International Conference for Learning Representations}.

\bibitem[{Kiros et~al.(2015)Kiros, Zhu, Salakhutdinov, Zemel, Urtasun,
  Torralba, and Fidler}]{kiros2015skip}
Ryan Kiros, Yukun Zhu, Ruslan~R Salakhutdinov, Richard Zemel, Raquel Urtasun,
  Antonio Torralba, and Sanja Fidler. 2015.
\newblock \href {https://papers.nips.cc/paper/5950-skip-thought-vectors.pdf}
  {Skip-thought vectors}.
\newblock In \emph{Advances in Neural Information Processing Systems}, pages
  3294--3302.

\bibitem[{Klein and Manning(2003)}]{klein2003accurate}
Dan Klein and Christopher~D Manning. 2003.
\newblock \href {https://www.aclweb.org/anthology/P03-1054.pdf} {Accurate
  unlexicalized parsing}.
\newblock In \emph{Proceedings of the 41st Annual Meeting on Association for
  Computational Linguistics-Volume 1}, pages 423--430. Association for
  Computational Linguistics.

\bibitem[{Lake and Baroni(2018)}]{lake2018generalization}
Brenden~M. Lake and Marco Baroni. 2018.
\newblock \href {https://arxiv.org/pdf/1711.00350.pdf} {Generalization without
  systematicity: On the compositional skills of sequence-to-sequence recurrent
  networks}.
\newblock In \emph{International Conference on Machine Learning}.

\bibitem[{Lakretz et~al.(2019)Lakretz, Kruszewski, Desbordes, Hupkes, Dehaene,
  and Baroni}]{lakretz2019emergence}
Yair Lakretz, German Kruszewski, Theo Desbordes, Dieuwke Hupkes, Stanislas
  Dehaene, and Marco Baroni. 2019.
\newblock \href {https://doi.org/10.18653/v1/N19-1002} {The emergence of number
  and syntax units in {LSTM} language models}.
\newblock In \emph{Proceedings of the 2019 Conference of the North {A}merican
  Chapter of the Association for Computational Linguistics: Human Language
  Technologies, Volume 1 (Long and Short Papers)}, pages 11--20, Minneapolis,
  Minnesota. Association for Computational Linguistics.

\bibitem[{Li et~al.(2019)Li, Zhao, Wang, and
  Hestness}]{li-etal-2019-compositional}
Yuanpeng Li, Liang Zhao, Jianyu Wang, and Joel Hestness. 2019.
\newblock \href {https://doi.org/10.18653/v1/D19-1438} {Compositional
  generalization for primitive substitutions}.
\newblock In \emph{Proceedings of the 2019 Conference on Empirical Methods in
  Natural Language Processing and the 9th International Joint Conference on
  Natural Language Processing (EMNLP-IJCNLP)}, pages 4293--4302, Hong Kong,
  China. Association for Computational Linguistics.

\bibitem[{Linzen et~al.(2016)Linzen, Dupoux, and
  Goldberg}]{linzen2016assessing}
Tal Linzen, Emmanuel Dupoux, and Yoav Goldberg. 2016.
\newblock \href
  {https://www.mitpressjournals.org/doi/pdfplus/10.1162/tacl_a_00115}
  {Assessing the ability of {LSTM}s to learn syntax-sensitive dependencies}.
\newblock \emph{Transactions of the ACL}.

\bibitem[{Marvin and Linzen(2018)}]{marvin2018targeted}
Rebecca Marvin and Tal Linzen. 2018.
\newblock \href {https://doi.org/10.18653/v1/D18-1151} {Targeted syntactic
  evaluation of language models}.
\newblock In \emph{Proceedings of the 2018 Conference on Empirical Methods in
  Natural Language Processing}, pages 1192--1202, Brussels, Belgium.
  Association for Computational Linguistics.

\bibitem[{McCoy et~al.(2019{\natexlab{a}})McCoy, Linzen, Dunbar, and
  Smolensky}]{mccoy}
R.~Thomas McCoy, Tal Linzen, Ewan Dunbar, and Paul Smolensky.
  2019{\natexlab{a}}.
\newblock \href {https://openreview.net/forum?id=BJx0sjC5FX} {{RNN}s implicitly
  implement tensor-product representations}.
\newblock In \emph{International Conference on Learning Representations}.

\bibitem[{McCoy et~al.(2019{\natexlab{b}})McCoy, Pavlick, and
  Linzen}]{mccoy2019right}
R.~Thomas McCoy, Ellie Pavlick, and Tal Linzen. 2019{\natexlab{b}}.
\newblock \href {https://doi.org/10.18653/v1/P19-1334} {Right for the wrong
  reasons: Diagnosing syntactic heuristics in natural language inference}.
\newblock In \emph{Proceedings of the 57th Annual Meeting of the Association
  for Computational Linguistics}, pages 3428--3448, Florence, Italy.
  Association for Computational Linguistics.

\bibitem[{Mikolov et~al.(2013)Mikolov, Yih, and Zweig}]{mikolov2013linguistic}
Tomas Mikolov, Wen-tau Yih, and Geoffrey Zweig. 2013.
\newblock \href {https://www.aclweb.org/anthology/N13-1090} {Linguistic
  regularities in continuous space word representations}.
\newblock In \emph{Proceedings of the 2013 Conference of the North {A}merican
  Chapter of the Association for Computational Linguistics: Human Language
  Technologies}, pages 746--751, Atlanta, Georgia. Association for
  Computational Linguistics.

\bibitem[{Mu and Andreas(2020)}]{mu2020compositional}
Jesse Mu and Jacob Andreas. 2020.
\newblock \href {https://arxiv.org/pdf/2006.14032.pdf} {Compositional
  explanations of neurons}.
\newblock In \emph{Advances in Neural Information Processing Systems 33}.

\bibitem[{Omlin and Giles(1996)}]{omlin1996extraction}
Christian~W Omlin and C~Lee Giles. 1996.
\newblock Extraction of rules from discrete-time recurrent neural networks.
\newblock \emph{Neural networks}, 9(1):41--52.

\bibitem[{Palangi et~al.(2017)Palangi, Smolensky, He, and Deng}]{palangi}
Hamid Palangi, Paul Smolensky, Xiaodong He, and Li~Deng. 2017.
\newblock \href {https://arxiv.org/pdf/1705.08432.pdf} {Question-answering with
  grammatically-interpretable representations}.
\newblock In \emph{Proceedings of the Association for the Advancement of
  Artificial Intelligence}.

\bibitem[{Parikh et~al.(2016)Parikh, T{\"a}ckstr{\"o}m, Das, and
  Uszkoreit}]{parikh2016decomposable}
Ankur Parikh, Oscar T{\"a}ckstr{\"o}m, Dipanjan Das, and Jakob Uszkoreit. 2016.
\newblock \href {https://doi.org/10.18653/v1/D16-1244} {A decomposable
  attention model for natural language inference}.
\newblock In \emph{Proceedings of the 2016 Conference on Empirical Methods in
  Natural Language Processing}, pages 2249--2255, Austin, Texas. Association
  for Computational Linguistics.

\bibitem[{Pearl(2000)}]{pearl2000causality}
Judea Pearl. 2000.
\newblock \emph{Causality}.
\newblock MIT Press, Cambridge, MA.

\bibitem[{Peters et~al.(2018)Peters, Neumann, Zettlemoyer, and
  Yih}]{peters2018dissecting}
Matthew Peters, Mark Neumann, Luke Zettlemoyer, and Wen-tau Yih. 2018.
\newblock \href {https://doi.org/10.18653/v1/D18-1179} {Dissecting contextual
  word embeddings: Architecture and representation}.
\newblock In \emph{Proceedings of the 2018 Conference on Empirical Methods in
  Natural Language Processing}, pages 1499--1509, Brussels, Belgium.
  Association for Computational Linguistics.

\bibitem[{Poliak et~al.(2018)Poliak, Haldar, Rudinger, Hu, Pavlick, White, and
  Van~Durme}]{poliak2018collecting}
Adam Poliak, Aparajita Haldar, Rachel Rudinger, J.~Edward Hu, Ellie Pavlick,
  Aaron~Steven White, and Benjamin Van~Durme. 2018.
\newblock \href {https://doi.org/10.18653/v1/D18-1007} {Collecting diverse
  natural language inference problems for sentence representation evaluation}.
\newblock In \emph{Proceedings of the 2018 Conference on Empirical Methods in
  Natural Language Processing}, pages 67--81, Brussels, Belgium. Association
  for Computational Linguistics.

\bibitem[{Ravichander et~al.(2020)Ravichander, Belinkov, and
  Hovy}]{ravichander2020probing}
Abhilasha Ravichander, Yonatan Belinkov, and Eduard Hovy. 2020.
\newblock Probing the probing paradigm: Does probing accuracy entail task
  relevance?
\newblock \emph{arXiv preprint arXiv:2005.00719}.

\bibitem[{Shen et~al.(2019)Shen, Tan, Sordoni, and Courville}]{shen2018ordered}
Yikang Shen, Shawn Tan, Alessandro Sordoni, and Aaron Courville. 2019.
\newblock \href {https://openreview.net/forum?id=B1l6qiR5F7} {Ordered neurons:
  Integrating tree structures into recurrent neural networks}.
\newblock In \emph{International Conference on Learning Representations}.

\bibitem[{Smolensky(1990)}]{Smolensky:1990:TPV:102418.102425}
Paul Smolensky. 1990.
\newblock \href {https://doi.org/10.1016/0004-3702(90)90007-M} {Tensor product
  variable binding and the representation of symbolic structures in
  connectionist systems}.
\newblock \emph{Artif. Intell.}, 46(1-2):159--216.

\bibitem[{Socher et~al.(2013)Socher, Perelygin, Wu, Chuang, Manning, Ng, and
  Potts}]{socher2013recursive}
Richard Socher, Alex Perelygin, Jean Wu, Jason Chuang, Christopher~D. Manning,
  Andrew Ng, and Christopher Potts. 2013.
\newblock \href {https://www.aclweb.org/anthology/D13-1170} {Recursive deep
  models for semantic compositionality over a sentiment treebank}.
\newblock In \emph{Proceedings of the 2013 Conference on Empirical Methods in
  Natural Language Processing}, pages 1631--1642, Seattle, Washington, USA.
  Association for Computational Linguistics.

\bibitem[{Sutskever et~al.(2014)Sutskever, Vinyals, and
  Le}]{sutskever2014sequence}
Ilya Sutskever, Oriol Vinyals, and Quoc~V. Le. 2014.
\newblock \href
  {https://papers.nips.cc/paper/5346-sequence-to-sequence-learning-with-neural-networks.pdf}
  {Sequence to sequence learning with neural networks}.
\newblock In \emph{Advances in Neural Information Processing Systems}, pages
  3104--3112.

\bibitem[{Tenney et~al.(2019)Tenney, Xia, Chen, Wang, Poliak, McCoy, Kim,
  Durme, Bowman, Das, and Pavlick}]{tenney2018what}
Ian Tenney, Patrick Xia, Berlin Chen, Alex Wang, Adam Poliak, R.~Thomas McCoy,
  Najoung Kim, Benjamin~Van Durme, Sam Bowman, Dipanjan Das, and Ellie Pavlick.
  2019.
\newblock \href {https://openreview.net/forum?id=SJzSgnRcKX} {What do you learn
  from context? probing for sentence structure in contextualized word
  representations}.
\newblock In \emph{International Conference on Learning Representations}.

\bibitem[{Vanmassenhove et~al.(2017)Vanmassenhove, Du, and
  Way}]{vanmassenhove2017investigating}
Eva Vanmassenhove, Jinhua Du, and Andy Way. 2017.
\newblock Investigating `aspect' in {NMT} and {SMT}: Translating the {English}
  simple past and present perfect.
\newblock \emph{Computational Linguistics in the Netherlands Journal},
  7:109--128.

\bibitem[{Vaswani et~al.(2017)Vaswani, Shazeer, Parmar, Uszkoreit, Jones,
  Gomez, Kaiser, and Polosukhin}]{vaswani2017attention}
Ashish Vaswani, Noam Shazeer, Niki Parmar, Jakob Uszkoreit, Llion Jones,
  Aidan~N Gomez, {\L}ukasz Kaiser, and Illia Polosukhin. 2017.
\newblock \href
  {https://papers.nips.cc/paper/7181-attention-is-all-you-need.pdf} {Attention
  is all you need}.
\newblock In \emph{Advances in Neural Information Processing Systems}, pages
  5998--6008.

\bibitem[{Voita and Titov(2020)}]{Voita2020InformationTheoreticPW}
Elena Voita and Ivan Titov. 2020.
\newblock \href {https://arxiv.org/abs/2003.12298} {Information-theoretic
  probing with minimum description length}.
\newblock \emph{arXiv preprint arXiv:2003.12298}.

\bibitem[{Wang et~al.(2018)Wang, Singh, Michael, Hill, Levy, and
  Bowman}]{wang2018glue}
Alex Wang, Amanpreet Singh, Julian Michael, Felix Hill, Omer Levy, and Samuel
  Bowman. 2018.
\newblock \href {https://doi.org/10.18653/v1/W18-5446} {{GLUE}: A multi-task
  benchmark and analysis platform for natural language understanding}.
\newblock In \emph{Proceedings of the 2018 {EMNLP} Workshop {B}lackbox{NLP}:
  Analyzing and Interpreting Neural Networks for {NLP}}, pages 353--355,
  Brussels, Belgium. Association for Computational Linguistics.

\bibitem[{Warstadt et~al.(2020)Warstadt, Parrish, Liu, Mohananey, Peng, Wang,
  and Bowman}]{warstadt2019blimp}
Alex Warstadt, Alicia Parrish, Haokun Liu, Anhad Mohananey, Wei Peng, Sheng-Fu
  Wang, and Samuel~R Bowman. 2020.
\newblock \href {https://scholarworks.umass.edu/scil/vol3/iss1/43/} {{BLiMP}: A
  benchmark of linguistic minimal pairs for english}.
\newblock \emph{Proceedings of the Society for Computation in Linguistics.}

\bibitem[{Weiss et~al.(2018)Weiss, Goldberg, and Yahav}]{weiss2018extracting}
Gail Weiss, Yoav Goldberg, and Eran Yahav. 2018.
\newblock \href {http://proceedings.mlr.press/v80/weiss18a.html} {Extracting
  automata from recurrent neural networks using queries and counterexamples}.
\newblock In \emph{{International Conference on Machine Learning}}, pages
  5244--5253.

\bibitem[{Wickelgren(1969)}]{wickelgren1969context}
Wayne~A. Wickelgren. 1969.
\newblock Context-sensitive coding, associative memory, and serial order in
  (speech) behavior.
\newblock \emph{Psychological Review}, 76(1):1--15.

\bibitem[{Wu et~al.(2016)Wu, Schuster, Chen, Le, Norouzi, Macherey, Krikun,
  Cao, Gao, Macherey, Klingner, Shah, Johnson, Liu, Kaiser, Gouws, Kato, Kudo,
  Kazawa, Stevens, Kurian, Patil, Wang, Young, Smith, Riesa, Rudnick, Vinyals,
  Corrado, Hughes, and Dean}]{googlenmt}
Yonghui Wu, Mike Schuster, Zhifeng Chen, Quoc~V. Le, Mohammad Norouzi, Wolfgang
  Macherey, Maxim Krikun, Yuan Cao, Qin Gao, Klaus Macherey, Jeff Klingner,
  Apurva Shah, Melvin Johnson, Xiaobing Liu, Lukasz Kaiser, Stephan Gouws,
  Yoshikiyo Kato, Taku Kudo, Hideto Kazawa, Keith Stevens, George Kurian,
  Nishant Patil, Wei Wang, Cliff Young, Jason Smith, Jason Riesa, Alex Rudnick,
  Oriol Vinyals, Greg Corrado, Macduff Hughes, and Jeffrey Dean. 2016.
\newblock \href {https://arxiv.org/abs/1609.08144} {Google's neural machine
  translation system: Bridging the gap between human and machine translation}.
\newblock \emph{arXiv preprint arXiv:1609.08144}.

\end{thebibliography}

\clearpage

\appendix
\section{Appendix}

\subsection{Designed role schemes} \label{sec:role-schemes}
We use six hand-specified role schemes as a baseline to compare the learned role schemes against. Examples of each role scheme are shown in Table \ref{tab:rolestwo}.

\begin{enumerate}
	\item Left-to-right (LTR): Each filler's role is its index in the sequence, counting from left to right.
	\item Right-to-left (RTL): Each filler's role is its index in the sequence, counting from right to left.
	\item Bidirectional (Bi): Each filler's role is a pair of indices, where the first index counts from left to right, and the second index counts from right to left.
	\item Tree: Each filler's role is given by its position in a tree. This depends on a tree parsing algorithm.
	\item Wickelroles (Wickel): Each filler's role is a 2-tuple containing the filler before it and the filler after it. \citep{wickelgren1969context}
	\item Bag-of-words (BOW): Each filler is assigned the same role. The position and context of the filler is ignored.
\end{enumerate}

\begin{table*}[ht]
    \centering
     \resizebox{\textwidth}{!}{
    \begin{tabular} {l|llll|llllll}
    \toprule
    & 3 & 1 & 1& 6 & 5 & 2 & 3 & 1 & 9 & 7 \\ \midrule
    Left-to-right & 0 &1&2&3 & 0&1&2&3&4&5\\
Right-to-left & 3&2&1&0 & 5&4&3&2&1&0 \\
Bidirectional & (0, 3)&(1, 2)&(2, 1)&(3, 0) & (0, 5)&(1, 4)&(2, 3)&(3, 2)&(4, 1)&(5, 0) \\
Wickelroles & \#\_1& 3\_1& 1\_6& 1\_\# & \#\_2& 5\_3& 2\_1& 3\_9& 1\_7& 9\_\# \\
Tree & L& RLL& RLR& RR & LL& LRLL& LRLR& LRRL& LRRR& R \\
Bag of words & r$_0$&r$_0$&r$_0$&r$_0$ & r$_0$&r$_0$&r$_0$&r$_0$&r$_0$&r$_0$ \\ \bottomrule
\end{tabular}
}
    \caption{\label{tab:rolestwo}The assigned roles for two sequences, 3116 and 523197. Table reproduced from \citet{mccoy}.}
\end{table*}

\subsection{ROLE regularization} \label{sec:role-regulatization}

Letting $\mA = \{\va_t \}_{t=1}^T$, the regularization term applied during ROLE training is $R = \lambda (R_1 + R_2 + R_3)$, where $\lambda$ is a regularization hyperparameter and:
\begin{align*} 
    R_1(\mA) &= \sum_{t=1}^T\sum_{\rho=1}^{n_\rR} [\va_t]_\rho(1-[\va_t]_\rho); \hspace{1.5mm}\\
    R_2(\mA) &= -\sum_{t=1}^{T}\sum_{\rho=1}^{n_\rR} [\va_t]_\rho^2; \hspace{1.5mm}\\
    R_3(\mA) &= \sum_{\rho=1}^{n_\rR} \left( [\vs_\mA]_\rho (1 - [\vs_\mA]_\rho) \right)^2
\end{align*}
Since each $\va_t$ results from a softmax, its elements are positive and sum to 1. Thus the factors in $R_1(\mA)$ are all non-negative, so $R_1$ assumes its minimal value of 0 when each $\va_t$ has binary elements; since these elements must sum to 1, such an $\va_t$ must be one-hot.
$R_2(\mA)$ is also minimized when each $\va_t$ is one-hot because when a vector's $L^1$ norm is 1, its $L^2$ norm is maximized when it is one-hot. Although each of these terms individually favor one-hot vectors, empirically we find that using both terms helps the training process.
In a discrete symbolic structure, each position can hold at most one symbol, and the final term $R_3$ in \RLN's regularizer $R$ is designed to encourage this.
In the vector $\vs_\mA = \sum_{t=1}^T \va_t$, the $\rho^\r{th}$ element is the total attention weight, over all symbols in the string, assigned to the $\rho^{\r{th}}$ role: in the discrete case, this must be 0 (if no symbol is assigned this role) or 1 (if a single symbol is assigned this role).
Thus $R_3$ is minimized when all elements of $\vs$ are 0 or 1 ($R_3$ is similar to $R_1$, but with squared terms since we are no longer assured each element is at most 1). It is important to normalize each role embedding in the role matrix \textbf{R} so that small attention weights have correspondingly small impacts on the weighted-sum role embedding.

\subsection{RNN trained on SCAN} \label{sec:rnn-scan}

To train the standard RNN on SCAN, we ran a limited hyperparameter search similar to the procedure in \citet{lake2018generalization}. Since our goal was to produce a single embedding that captured the entire input sequence, we fixed the architecture as a GRU with a single hidden layer. We did not train models with attention, to investigate whether a standard RNN could capture compositionality in its single bottleneck encoding. The remaining hyperparameters were hidden dimension and dropout. We ran a search over the hidden dimension sizes of 50, 100, 200, and 400 as well as dropout with a value of 0, .1, and .5 applied to the word embeddings and recurrent layer. Each network was trained with the Adam optimizer \citep{kingma2015adam} and a learning rate of .001 for 100,000 steps with a batch-size of 1. The best performing network had a hidden dimension or 100 and dropout of .1.

\subsection{\RLN\ trained on SCAN} \label{sec:role-scan}

For the \RLN\ models trained to approximate the GRU encoder trained on SCAN, we used a filler dimension of 100, and a role dimension of 50 with 50 roles available. For training, we used the Adam \citep{kingma2015adam} optimizer with a learning rate of .001, batch size 32, and an early stopping patience of 10. The role assignment module used a bidirectional 2-layer LSTM \citep{Hochreiter:1997:LSM:1246443.1246450}. We performed a hyperparameter search over the regularization coefficient $\lambda$ using the values in the set [.1, .02, .01]. The best performing value was .02, and we used this model in our analysis.


The algorithm below characterizes our post-hoc interpretation of which roles the Role Learner will assign to elements of the input to the SCAN model. This algorithm was created by hand based on an analysis of the Role Learner's outputs for the elements of the SCAN training set. The algorithm works equally well on examples in the training set and the test set; on both datasets, it exactly matches the roles chosen by the Role Learner for 98.7\% of sequences (20,642 out of 20,910).

\subsection{A role-assignment algorithm implicitly learned by the SCAN seq2seq encoder} \label{sec:Alg}

The input sequences have three basic types that are relevant to determining the role assignment: sequences that contain \textit{and} (e.g., \textit{jump around left and walk thrice}), sequences that contain  \textit{after} (e.g., \textit{jump around left after walk thrice}), and sequences without \textit{and} or \textit{after} (e.g., \textit{turn opposite right thrice}). Within commands containing \textit{and} or \textit{after}, it is convenient to break the command down into the command before the connecting word and the command after it; for example, in the command \textit{jump around left after walk thrice}, these two components would be \textit{jump around left} and \textit{walk thrice}. 

\begin{itemize}
    \item Sequence with \textit{and}:
    \begin{itemize}
        \item Elements of the command before \textit{and}:
        \begin{itemize}
            \item Last word: 28
            \item First word (if not also last word): 46
            \item \textit{opposite} if the command ends with \textit{thrice}: 22
            \item Direction word between \textit{opposite} and \textit{thrice}: 2
            \item \textit{opposite} if the command does not end with \textit{thrice}: 2
            \item Direction word after \textit{opposite} but not before \textit{thrice}: 4
            \item \textit{around}: 22
            \item Direction word after \textit{around}: 2
            \item Direction word between an action word and \textit{twice} or \textit{thrice}: 2
        \end{itemize}
        \item Elements of the command before \textit{and}:
        \begin{itemize}
            \item First word: 11
            \item Last word (if not also the first word): 36
            \item Second-to-last word (if not also the first word): 3
            \item Second of four words: 24
        \end{itemize}
        \item \textit{and}: 30
    \end{itemize}
    \item Sequence with \textit{after}:
    \begin{itemize}
        \item Elements of the command before \textit{after}:
        \begin{itemize}
            \item Last word: 8
            \item Second-to-last word: 36
            \item First word (if not the last or second-to-last word): 11
            \item Second word (if not the last or second-to-last word): 3
        \end{itemize}
        \item Elements of the command after \textit{after}:
        \begin{itemize}
            \item Last word: 46
            \item Second-to-last word: 4
            \item First word if the command ends with \textit{around right}: 4
            \item First word if the command ends with \textit{thrice} and contains a rotation: 10
            \item First word if the command does not end with \textit{around right} and does not contain both \textit{thrice} and a rotation: 17
            \item Second word if the command ends with \textit{thrice}: 17
            \item Second word if the command does not end with \textit{thrice}: 10
        \end{itemize}
        \item \textit{after}: 17 if no other word has role 17 or if the command after \textit{after} ends with \textit{around left}; 43 otherwise
    \end{itemize}
    \item Sequence without \textit{and} or \textit{after}:
    \begin{itemize}
        \item Action word directly before a cardinality: 4
        \item Action word before, but not directly before, a cardinality: 34
        \item \textit{thrice} directly after an action word: 2
        \item \textit{twice} directly after an action word: 2
        \item \textit{opposite} in a sequence ending with \textit{twice}: 8
        \item \textit{opposite} in a sequence ending with \textit{thrice}: 34
        \item \textit{around} in a sequence ending with a cardinality:  22
        \item Direction word directly before a cardinality: 2
        \item Action word in a sequence without a cardinality: 46
        \item \textit{opposite} in a sequence without a cardinality: 2
        \item Direction after \textit{opposite} in a sequence without a cardinality: 26
        \item \textit{around} in a sequence without a cardinality: 3
        \item Direction after \textit{around} in a sequence without a cardinality: 22
        \item Direction directly after an action in a sequence without a cardinality: 22
    \end{itemize}
\end{itemize}

\noindent
To show how this works with an example, consider the input \textit{jump around left after walk thrice}. The command before \textit{after} is \textit{jump around left}. \textit{left}, as the last word, is given role 8. \textit{around}, as the second-to-last word, gets role 36. \textit{jump}, as a first word that is not also the last or second-to-last word gets role 11. The command after \textit{after} is \textit{walk thrice}. \textit{thrice}, as the last word, gets role 46. \textit{walk}, as the second-to-last word, gets role 4. Finally, \textit{after} gets role 17 because no other elements have been assigned role 17 yet. These predicted outputs match those given by the Role Learner.

\subsection{Discussion of the algorithm} \label{sec:AlgDisc}

We offer several observations about this algorithm.

\begin{enumerate}
    \item 
This algorithm may seem convoluted, but a few observations can illuminate how the roles assigned by such an algorithm support success on the SCAN task. First, a sequence will contain role 30 if and only if it contains \textit{and}, and it will contain role 17 if and only if it contains \textit{after}. Thus, by implicitly checking for the presence of these two roles (regardless of the fillers bound to them), the decoder can tell whether the output involves one or two basic commands, where the presence of \textit{and} or \textit{after} leads to two basic commands and the absence of both leads to one basic command. Moreover, if there are two basic commands, whether it is role 17 or role 30 that is present can tell the decoder whether the input order of these commands also corresponds to their output order (when it is \textit{and} in play, i.e., role 30), or if the input order is reversed (when it is \textit{after} in play, i.e., role 17).

With these basic structural facts established, the decoder can begin to decode the specific commands. For example, if the input is a sequence with \textit{after}, it can begin with the command after \textit{after}, which it can decode by checking which fillers are bound to the relevant roles for that type of command.

It may seem odd that so many of the roles are based on position (e.g., ``first word" and ``second-to-last word"), rather than more functionally-relevant categories such as ``direction word." However, this approach may actually be more efficient: Each command consists of a single mandatory element (namely, an action word such as \textit{walk} or \textit{jump}) followed by several optional modifiers (namely, rotation words, direction words, and cardinalities). Because most of the word categories are optional, it might be inefficient to check for the presence of, e.g., a cardinality, since many sequences will not have one. By contrast, every sequence will have a last word, and checking the identity of the last word provides much functionally-relevant information: if that word is not a cardinality, then the decoder knows that there is no cardinality present in the command (because if there were, it would be the last word); and if it is a cardinality, then that is important to know, because the presence of \textit{twice} or \textit{thrice} can dramatically affect the shape of the output sequence. In this light, it is unsurprising that the SCAN encoder has implicitly learned several different roles that essentially mean the last element of a particular subcommand.
\item
The algorithm does not constitute a simple, transparent role scheme. But its job is to describe the representations that the original network produces, and we have no a priori expectation about how complex that process may be. The role-assignment algorithm implicitly learned by ROLE is interpretable locally (each line is readily expressible in simple English), but not intuitively transparent globally. We see this as a positive result, in two respects.

First, it shows why ROLE is crucial: no human-generated role scheme would provide a good approximation to this algorithm. Such an algorithm can only be identified because ROLE is able to use gradient descent to find role schemes far more complex than any we would hypothesize intuitively. This enables us to analyze networks far more complex than we could analyze previously, being necessarily limited to hand-designed role schemes based on human intuitions about how to perform the task. 

Second, when future work illuminates the computation in the original SCAN GRU seq2seq decoder, the baroqueness of the role-assignment algorithm that ROLE has shown to be implicit in the seq2seq encoder can potentially explain certain limitations in the original model, which is known to suffer from severe failures of systematic generalization outside the training distribution (Lake and Baroni, 2018). It is reasonable to hypothesize that systematic generalization requires that the encoder learn an implicit role scheme that is relatively simple and highly compositional. Future proposals for improving the systematic generalization of models on SCAN can be examined using ROLE to test the hypothesis that greater systematicity requires greater compositional simplicity in the role scheme implicitly learned by the encoder.

\item
While the role-assignment algorithm of A.8.1 may not be simple, from a certain perspective, 
it is quite surprising that it is not far more complex. Although ROLE is provided 50 roles to learn to deploy as it likes, it only chooses to use 16 of them (only 16 are ever selected as the $\argmax(\va_t)$; see Sec. 6.1). Furthermore, the SCAN grammar generates 20,910 input sequences, containing a total of 151,688 words (an average of 7.25 words per input). This means that, if one were to generate a series of conditional statements to determine which role is assigned to each word in every context, this could in theory require up to 151,688 conditionals (e.g., ``if the filler is `jump’ in the context `walk thrice after \underline{\hspace{.25 in}} opposite left’, then assign role 17''). However, our algorithm involves just 47 conditionals. This reduction helps explain how the model performs so well on the test set: If it used many more of the 151,688 possible conditional rules, it would completely overfit the training examples in a way that would be unlikely to generalize. The 47-conditional algorithm we found is more likely to generalize by abstracting over many details of the context.

\item
Were it not for ROLE’s ability to characterize the representations generated by the original encoder in terms of implicit roles, providing an equally complete and accurate interpretation of those representations would necessarily require identifying the conditions determining the activation level of each of the 100 neurons hosting those representations. It seems to us grossly overly optimistic to estimate that each neuron’s activation level in the representation of a given input could be characterized by a property of the input statable in, say, two lines of roughly 20 words/symbols; yet even then, the algorithm would require 200 lines, whereas the algorithm in A.8.1 requires 47 lines of that scale. Thus, by even such a crude estimate of the degree of complexity expected for an algorithm describing the representations in terms of neuron activities, the algorithm we find, stated over roles, is 4 times simpler.

\end{enumerate}

\subsection{TPEs trained on sentence embedding models} \label{sec:tpe-sentences}

For each sentence embedding model, we trained three randomly initialized TPEs for each role scheme and selected the best performing one as measured by the lowest MSE. For each TPE, we used the original filler embedding from the sentence embedding model. This filler dimensionality is 25 for SST, 300 for SPINN and InferSent, and 620 for Skipthought. We applied a linear transformation to the pre-trained filler embedding where the input size is the dimensionality of the pre-trained embedding and the output size is also the dimensionality of the pre-trained embedding. This linearly transformed embedding is used as the filler vector in the filler-role binding in the TPE. For each TPE, we use a role dimension of 50. Training was done with a batch size of 32 using the Adam optimizer with a learning rate of .001.

To generate tree roles from the English sentences, we used the constituency parser released in version 3.9.1 of Stanford CoreNLP \citep{klein2003accurate}.

\subsection{\RLN\ trained on sentence embedding models} \label{sec:role-sentences}

For each sentence embedding model, we trained three randomly initialized \RLN\ models and selected the best performing one as measured by the lowest MSE. We used the original filler embedding from the sentence embedding model (25 for SST, 300 for SPINN and InferSent, and 620 for Skipthought). We applied a linear transformation to the pre-trained filler embedding where the input size is the dimensionality of the pre-trained embedding and the output size is also the dimensionality of the pre-trained embedding. This linearly transformed embedding is used as the filler vector in the filler-role binding in the TPE. We also applied a similar linear transformation to the pre-trained filler embedding before input to the role learner LSTM. For each \RLN\ model, we provide up to 50 roles with a role dimension of 50. Training was done with a batch size of 32 using the ADAM optimizer with a learning rate of .001. We performed a hyperparameter search over the regularization coefficient $\lambda$ using the values in the set $\{1, 0.1, 0.01, 0.001, 0.0001 \}$. For SST, SPINN, InferSent and SST, respectively, the best performing network used $\lambda=0.001, 0.01, 0.001, 0.1$. 

\end{document}